\documentclass[10pt,twocolumn,letterpaper]{article}
\usepackage[pagenumbers]{cvpr} 
\usepackage[dvipsnames]{xcolor}
\usepackage{cite}
\usepackage{amsmath,amssymb,amsfonts}
\usepackage{algorithmic}
\usepackage{graphicx}
\usepackage{textcomp}
\usepackage{xcolor}
\usepackage{color}
\usepackage{colortbl}
\usepackage{mathtools}
\usepackage{amsmath}
\usepackage{tabularx}
\usepackage{multirow}
\usepackage{multicol}
\usepackage{float}
\usepackage{tabularx}
\usepackage{threeparttable}
\usepackage{booktabs}
\usepackage{longtable}
\usepackage{makecell}
\usepackage{verbatim}
\usepackage{wrapfig}
\usepackage{xspace}
\usepackage{pbox}
\usepackage{epstopdf}
\usepackage{comment}
\usepackage{lipsum}
\usepackage{bm}
\usepackage{bbm}
\usepackage{setspace}
\usepackage{sidecap}
\usepackage{array}
\usepackage{blindtext}
\usepackage{enumitem}
\usepackage{diagbox}
\usepackage{url}
\usepackage{indentfirst}
\definecolor{cvprblue}{rgb}{0.21,0.49,0.74}
\usepackage[pagebackref,breaklinks,colorlinks,citecolor=cvprblue]{hyperref}

\usepackage{svg}

\usepackage[capitalize]{cleveref}
\crefname{section}{Sec.}{Secs.}
\crefname{section}{Section}{Sections}
\crefname{table}{Table}{Tables}
\crefname{table}{Tab.}{Tabs.}

\newcommand{\dataset}{TACO\xspace}

\def\paperID{106} 
\def\confName{CVPR}
\def\confYear{2024}

\title{TACO: Benchmarking Generalizable Bimanual\\Tool-ACtion-Object Understanding}

\author{
Yun Liu\textsuperscript{1,2,3},~
Haolin Yang\textsuperscript{4},~
Xu Si\textsuperscript{1},~
Ling Liu\textsuperscript{5},~
Zipeng Li\textsuperscript{1},~
Yuxiang Zhang\textsuperscript{1},~
Yebin Liu\textsuperscript{1},~
Li Yi\textsuperscript{\textdagger,1,2,3}
\smallskip\\
\textsuperscript{1}Tsinghua University~~~
\textsuperscript{2}Shanghai Artificial Intelligence Laboratory~~~
\textsuperscript{3}Shanghai Qi Zhi Institute\\
\textsuperscript{4}Beijing University of Posts and Telecommunications~~~
\textsuperscript{5}Beijing Institute of Technology
}

\begin{document}

\twocolumn[{
\renewcommand\twocolumn[1][]{#1}
\maketitle
\begin{center}
    \captionsetup{type=figure}
    \includegraphics[width=1.0\textwidth]{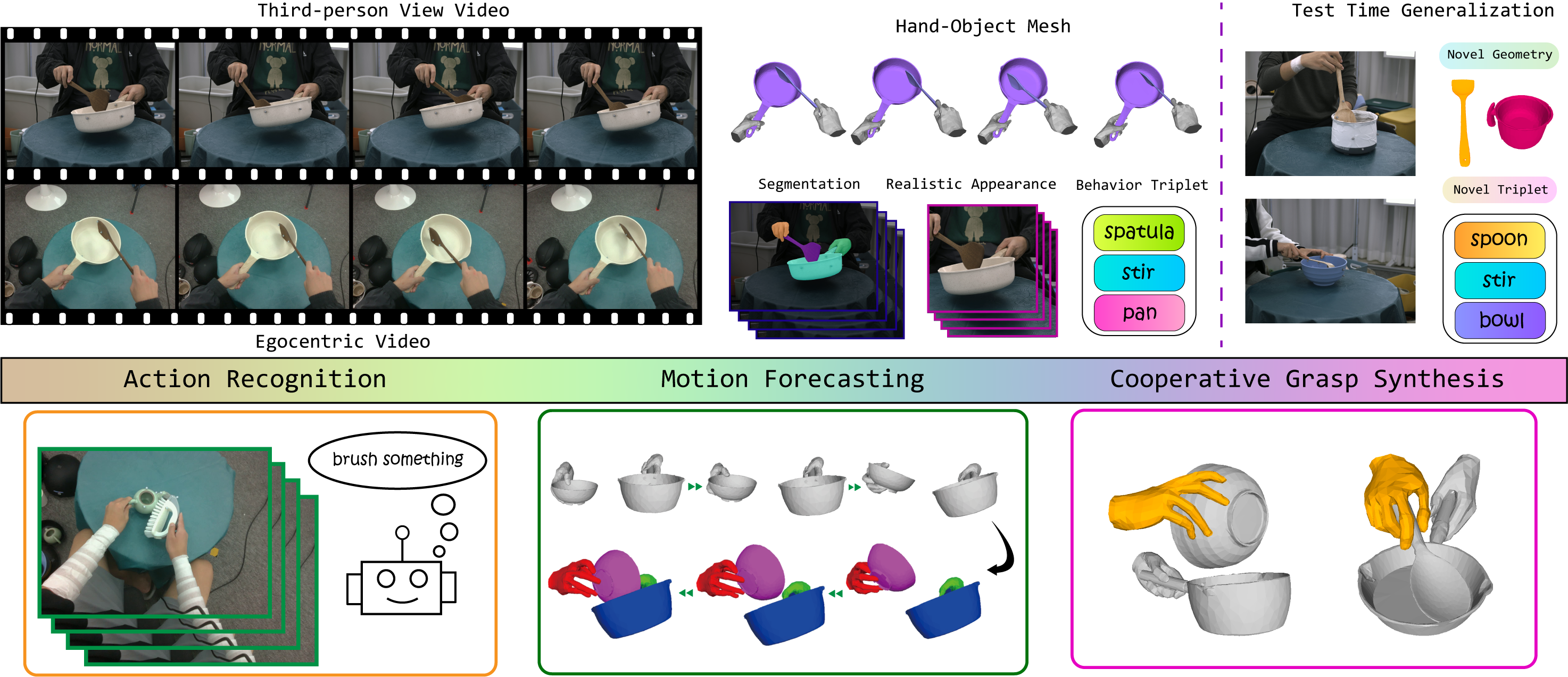}
    \caption{\dataset is a large-scale bimanual hand-object manipulation dataset covering extensive tool-action-object combinations in real-world scenarios. It involves videos from both 12 third-person views and one egocentric view, together with precise hand-object meshes, 2D segmentation, realistic hand-object appearances, and behavior triplet annotations. With rich diversities of object shapes and interaction behaviors, \dataset supports test-time generalization to unseen object geometries and novel behavior triplets and benchmarks various generalizable research topics, \textit{$e.g.$}, action recognition, motion forecasting, and cooperative grasp synthesis.}
    \label{fig:teaser}
\end{center}
}]

\footnotetext[2]{Corresponding author.}

\begin{abstract}
Humans commonly work with multiple objects in daily life and can intuitively transfer manipulation skills to novel objects by understanding object functional regularities. However, existing technical approaches for analyzing and synthesizing hand-object manipulation are mostly limited to handling a single hand and object due to the lack of data support. To address this, we construct \dataset, an extensive bimanual hand-object-interaction dataset spanning a large variety of tool-action-object compositions for daily human activities. \dataset contains 2.5K motion sequences paired with third-person and egocentric views, precise hand-object 3D meshes, and action labels. To rapidly expand the data scale, we present a fully automatic data acquisition pipeline combining multi-view sensing with an optical motion capture system. With the vast research fields provided by \dataset, we benchmark three generalizable hand-object-interaction tasks: compositional action recognition, generalizable hand-object motion forecasting, and cooperative grasp synthesis. Extensive experiments reveal new insights, challenges, and opportunities for advancing the studies of generalizable hand-object motion analysis and synthesis. Our data and code are available at \url{https://taco2024.github.io}.
\end{abstract}

\section{Introduction}

In our everyday lives, humans effortlessly synchronize the movements of both hands to manipulate a pair of objects, such as using a spatula to stir in a pan while cooking. Usually, these objects are not symmetric in their roles, with one acting as a tool to enhance the bimanual action performed on the other object. This allows us to characterize the bimanual behaviors with tool-action-object triplets. Various triplet combinations usually incorporate intricate and differing temporal and spatial coordination among the tool, the object, and the two hands. Understanding versatile bimanual behaviors could immediately benefit numerous applications in VR/AR, human-robot interaction~\cite{ng2023takes,christen2023learning}, and dexterous manipulation~\cite{D-Grasp,qin2022dexmv,CGF,zhang2023artigrasp}, which pose significant challenges for today’s computer vision systems. 

Tackling the above challenges requires the support of large-scale annotation-rich datasets. Existing dataset efforts~\cite{fhpa,brahmbhatt2020contactpose,hampali2020honnotate,chao2021dexycb,yang2022oakink,liu2022hoi4d,jian2023affordpose} on hand-object interactions (HOI) primarily focus on unimanual actions. However, simply aggregating two unimanual actions falls short of encompassing bimanual coordination behaviors. Furthermore, some works~\cite{kwon2021h2o,h2o3d,ye2021h2o_handover,xie2023hmdo,fan2023arctic} examine bimanual manipulation of a single object which can hardly extend to interpreting tool-action-object triplets. The most relevant dataset~\cite{krebs2021kit} currently available covers only 12 objects and fewer than 20 distinct tool-action-object triplets, severely limiting its potential to facilitate and benchmark the understanding of bimanual tool-action-object interactions that can generalize to novel object geometries or previously unseen triplets. This limitation partially stems from the challenging nature of jointly capturing two dynamic hands interacting with two objects in the real world, but further underscores the urgency of developing methodologies for generalizable bimanual HOI understanding. Therefore, we put generalization as our primary focus while exploring the curation of a large-scale bimanual HOI dataset that can both facilitate and benchmark generalizable understanding.

In particular, we present \textbf{\dataset},
a large-scale bimanual manipulation dataset encompassing a diverse array of tool-action-object compositions in real-world settings. Serving as a knowledge base of multi-object cooperation, \dataset focuses on daily scenarios involving the use of tools to interact with target objects and collects 131 types of $<$tool category, action label, target object category$>$ triplets across 20 object categories, 196 object instances, and 15 daily actions. We organize the dataset based on the triplets and make sure triplets have different levels of overlaps, this naturally defines the semantic distance between different motion trajectories and supports studying generalization with different extents. To expedite the expansion of our dataset, we combine the advantages of marker-based and markerless motion capture systems and present a fully automatic data acquisition pipeline that can guarantee motion quality and visual data quality at the same time. For each time step, the pipeline automatically provides labels including the precise recovery of hand-object mesh and segmentation on markerless vision data. As a result, \dataset comprises a total of 2.5K motion sequences and 5.2M video frames incorporating 3rd-person and egocentric views.

Benefiting from the rich and diverse motion sequences within our dataset, we carefully benchmark three tasks aiming at generalizing common human manipulation knowledge to unseen object geometries and novel tool-action-object combinations: \textbf{1)} compositional action recognition, \textbf{2)} generalizable hand-object motion forecasting, and \textbf{3)} cooperative grasp synthesis. Extensive studies are conducted on these benchmarks: First, diverse hand-object interactive motions with action annotations pave the way to learning to recognize hand-object action under novel object combinations in real-world scenarios. Additionally, the availability of 4D dynamic hand-object pose sequences allows for fine-grained hand-object motion tendency forecasting. Finally, we provide contact-rich hand-object meshes to facilitate studying hand grasp synthesis for unseen object geometries and categories under interaction scenarios.
Extensive experiments have been conducted on these benchmarks, revealing generalization challenges faced by existing methodologies when dealing with novel object geometries, object categories, and interaction triplets. We anticipate that \dataset will offer more opportunities for researching and developing universal generalization strategies in the understanding and creation of hand-object interactions.

In summary, our main contributions are threefold:

\begin{itemize}

\item We present \dataset, the first large-scale real-world 4D bimanual hand-object-interaction dataset covering diverse tool-action-object compositions and object geometries.

\item We develop an automatic data acquisition pipeline that provides precise recovery of hand-object mesh and segmentation together with markerless vision data.

\item We benchmark three tasks toward generalizable hand-object motion analysis and synthesis. We provide a comprehensive discussion and highlight the new challenges posed by \dataset.

\end{itemize}

\section{Related Work}
\subsection{Hand-object Interaction Datasets}

Understanding hand-object interaction is an emerging research topic that is supported by a large number of datasets. Many widely-used video datasets~\cite{epic-kitchens,kit_bimanual_actions,ego4d,sener2022assembly101} capture 2D real-world hand-object interaction videos to facilitate traditional visual perception studies such as action recognition and HOI detection. With the rapid development of 3D computer vision, recent advances have incorporated 3D hand-object mesh annotations into data-capturing processes and presented various 3D~\cite{obman,jian2023affordpose} or 4D~\cite{fhpa,brahmbhatt2020contactpose,taheri2020grab,hampali2020honnotate,kit_bimanual_actions,kwon2021h2o,chao2021dexycb,h2o3d,ye2021h2o_handover,yang2022oakink,liu2022hoi4d,xie2023hmdo,fan2023arctic} datasets with different focuses. Table \ref{tab:dataset} briefly summarizes the characteristics of these datasets. A majority of existing datasets focus on single-hand grasping, a basic manipulation skill that can benefit existing dexterous robot learning~\cite{qin2022dexmv,D-Grasp,qin2022one,zhang2023artigrasp}. Beyond grasping, a line of datasets captures diverse and complex manipulation skills requiring humans to perform complicated, multi-step actions like using tools~\cite{krebs2021kit,liu2022hoi4d,fan2023arctic}, deforming soft bodies~\cite{xie2023hmdo}, and multi-person handovers~\cite{ye2021h2o_handover}. From the perspective of object behaviors, KIT~\cite{krebs2021kit}, HOI4D~\cite{liu2022hoi4d}, and AffordPose\cite{jian2023affordpose} capture functional manipulations aiming at showing actual functions and usages of objects. KIT further collects multi-object interaction behaviors, however, it lacks object diversities and is hard to support generalizable studies. Our dataset captures diverse bimanual functional manipulation behaviors for multi-object cooperation, serving a wide range of research directions.

\begin{table*}[h!]
\centering
\scriptsize
\addtolength{\tabcolsep}{-3pt}
{
\begin{tabular}{c|ccccccccc|cc}
\toprule 
\multirow{3}{*}{dataset} & \multicolumn{9}{c|}{data characteristics:} & \multicolumn{2}{c}{\# number of:} \\
 & bimanual & multi-object & functional & category-level & egocentric & multi-view & markerless & mocap & action & sequence & frame \\
 & & cooperation & manipulation & & & & & & label & & \\
\midrule
FHPA~\cite{fhpa} & & & & & \checkmark & & & \checkmark & \checkmark & 1.2K & 105K \\
ObMan~\cite{obman} & & & & & & & & & & - & 154K \\
ContactPose\cite{brahmbhatt2020contactpose} & & & & & & \checkmark & & \checkmark & & 2.3K & 3.0M \\
GRAB~\cite{taheri2020grab} & \checkmark & & & & & & & \checkmark & & 1.3K & 1.6M \\
HO3D~\cite{hampali2020honnotate} & & & & & & \checkmark & \checkmark & & & 27 & 78K \\
KIT Binamual Manipulation~\cite{krebs2021kit} & \checkmark & \checkmark & \checkmark & & \checkmark &\checkmark & & \checkmark & \checkmark & 588 & 1.6M \\
H2O~\cite{kwon2021h2o} & \checkmark & & & & \checkmark & \checkmark & \checkmark & & & 191 & 571K \\
DexYCB~\cite{chao2021dexycb} & & & & & & \checkmark & \checkmark & & & 1.0K & 582K \\
H2O-3D~\cite{h2o3d} & \checkmark & & & & & \checkmark & \checkmark & & & 17 & 76K \\
H2O for handover~\cite{ye2021h2o_handover} & \checkmark & & & & & \checkmark & & \checkmark & & 6.0K & 5.0M \\
OakInk~\cite{yang2022oakink} & & & & & & \checkmark & & \checkmark & & 778 & 314K \\
HOI4D~\cite{liu2022hoi4d} & & & \checkmark & \checkmark & \checkmark & & \checkmark & & \checkmark & 4.0K & 1.2M \\
HMDO~\cite{xie2023hmdo} & \checkmark & & & & & \checkmark & \checkmark & & & 12 & 21K \\
ARCTIC~\cite{fan2023arctic} & \checkmark & & & & \checkmark & \checkmark & & \checkmark & & 339 & 2.1M \\
AffordPose~\cite{jian2023affordpose} & & & \checkmark & \checkmark & & & & & \checkmark & - & 27K \\
\hline
\dataset (ours) & \checkmark & \checkmark & \checkmark & \checkmark & \checkmark & \checkmark & \checkmark & \checkmark & \checkmark & 2.5K & 5.2M \\
\bottomrule
\end{tabular}
}
\vspace{-0.3cm}
\caption{Comparison of \dataset with existing 3D hand-object interaction datasets.}
\vspace{-0.5cm}
\label{tab:dataset}
\end{table*}

\subsection{Generalizable Hand-object Motion Analysis}

Generalizable hand-object-interaction motion analysis aims to understand the attributes of hand-object motion when manipulating novel object instances. To precisely recognize hand actions during the manipulation of unseen objects, a series of compositional action recognition approaches~\cite{materzynska2020something,kim2021motion,sun2021counterfactual,radevski2021revisiting,herzig2022object,yan2022look,zhang2022object,rajendiran2023modelling} represent objects through their bounding boxes to prevent the network from overfitting to intricate object geometries. To detect and infer action-object pairs under few-shot or zero-shot settings, a prevailing methodology used in recent generalizable HOI detection methods~\cite{hou2020visual,hou2021detecting,hou2021affordance,wang2022learning,zhou2022human,wang2023detecting} decomposes actions and objects into distinct features. This decomposition allows leveraging abundant training data with either similar actions or similar objects to handle rare action-object pairs that emerge at test time. It also provides the potential to discover novel and reasonable action-object pairs~\cite{hou2022discovering}. To estimate hand and object poses without object geometries, Zhu \textit{et al.}~\cite{ContactArt} propose learning hand-object interaction priors for each object category and transferring them to unseen objects within the same category during test time, enhancing the method adaptability to diverse unseen object instances.

\subsection{Generalizable Hand-object Motion Synthesis}

Synthesizing realistic hand-object-interaction motion on novel object geometries and categories is an emerging research topic that includes two main challenges: how to model generic, realistic, and diverse hand motion patterns from existing HOI data, and how to apply them to novel object instances. To address the first challenge, a branch of data-driven approaches decomposes complicated hand-object contact into spatial relations between each \textit{hand joint}~\cite{HALO,zheng2023cams,ContactGen,DexRepNet} or \textit{hand surface point}~\cite{CPF,manipnet,TOCH} and the object's relevant closest local regions, and such relations can be formulated as a combination of whether contact occurs~\cite{CPF,manipnet,TOCH,zheng2023cams,ContactGen,DexRepNet}, hand-object distances~\cite{CPF,TOCH,manipnet,DexRepNet}, positions~\cite{manipnet,TOCH,zheng2023cams,ContactGen,DexRepNet} and directions~\cite{zheng2023cams,ContactGen,DexRepNet} of contact points, and whether contact points should be closer~\cite{CPF}. Another line of work~\cite{grasping_field,HALO,grasp_tta,Contact2Grasp} encodes hand-object interaction integrally to an implicit global feature by a conditional variational auto-encoder (CVAE)~\cite{CVAE} structure. As solutions to the second challenge, generalizable methods~\cite{HALO,grasp_tta,zheng2023cams,ContactGen,Contact2Grasp} decompose the object's geometry from HOI and encode it as a condition for generative models. CAMS~\cite{zheng2023cams} further maps objects into a canonicalized object space to decrease the gap in object geometry. One limitation of existing approaches is that they do not focus on multi-hand and multi-object cooperation due to the lack of data support.

\section{Constructing \dataset}
In this section, we respectively describe the data capturing system (Section \ref{sec:3.1}), the data annotating pipeline (Section \ref{sec:3.2}), and dataset statistics (Section \ref{sec:3.3}).

\subsection{Data Capturing}
\label{sec:3.1}

\begin{figure}[h!]
    \centering
    \includegraphics[width=1.0\linewidth]{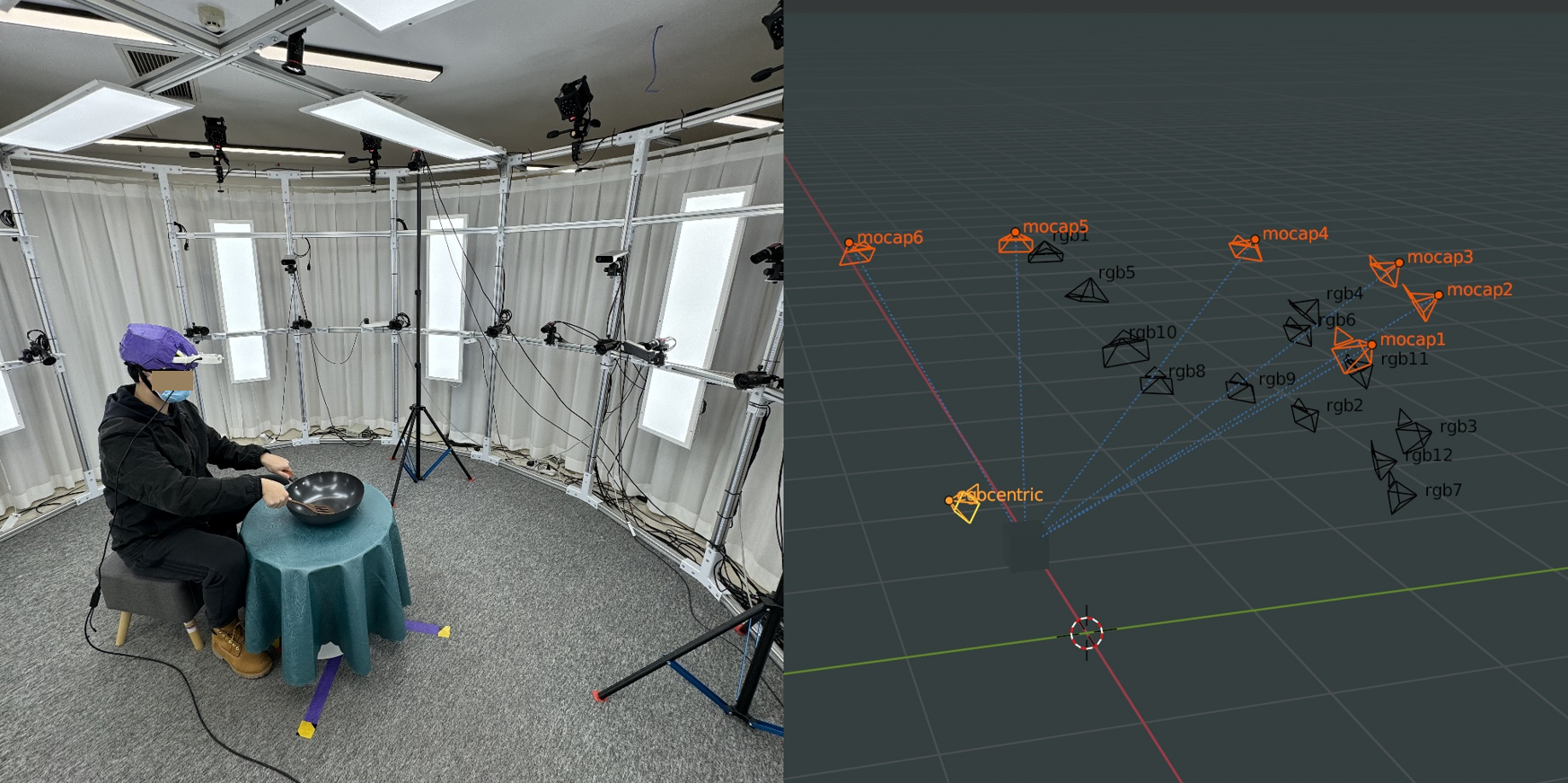}
    \vspace{-0.5cm}
    \caption{Data capturing system and camera views.}
    \vspace{-0.3cm}
   \label{fig:data_capturing_system}
\end{figure}


We obtain hand motion from multi-view RGB videos while capturing the object motion by attaching four markers on the object surface and tracking them with a mocap system. To this end, as shown in Figure \ref{fig:data_capturing_system}, our data capturing system incorporates 12 synchronized industrial FLIR cameras and a NOKOV optical motion capture suite with 6 infrared Mars4H cameras. To capture egocentric RGBD videos, a helmet equipped with a Realsense L515 camera is worn by the actor. Similar to objects, the motion of the L515 camera is tracked by the mocap system. The frequency of our cameras and mocap system is 30 Hz. The resolutions of our allocentric and egocentric images are 4096x3000 and 1920x1080, respectively.

\textbf{Object model acquisition.} To capture contacts between hands and objects and support relevant studies, we obtain an accurate 3D mesh for each object with an industrial EinScan 3D scanner. Each object mesh is represented by up to 100K triangular faces for fine-grained geometries.

\subsection{Data Annotating}
\label{sec:3.2}

\begin{figure*}[h!]
  \centering
   \includegraphics[width=1.0\linewidth]{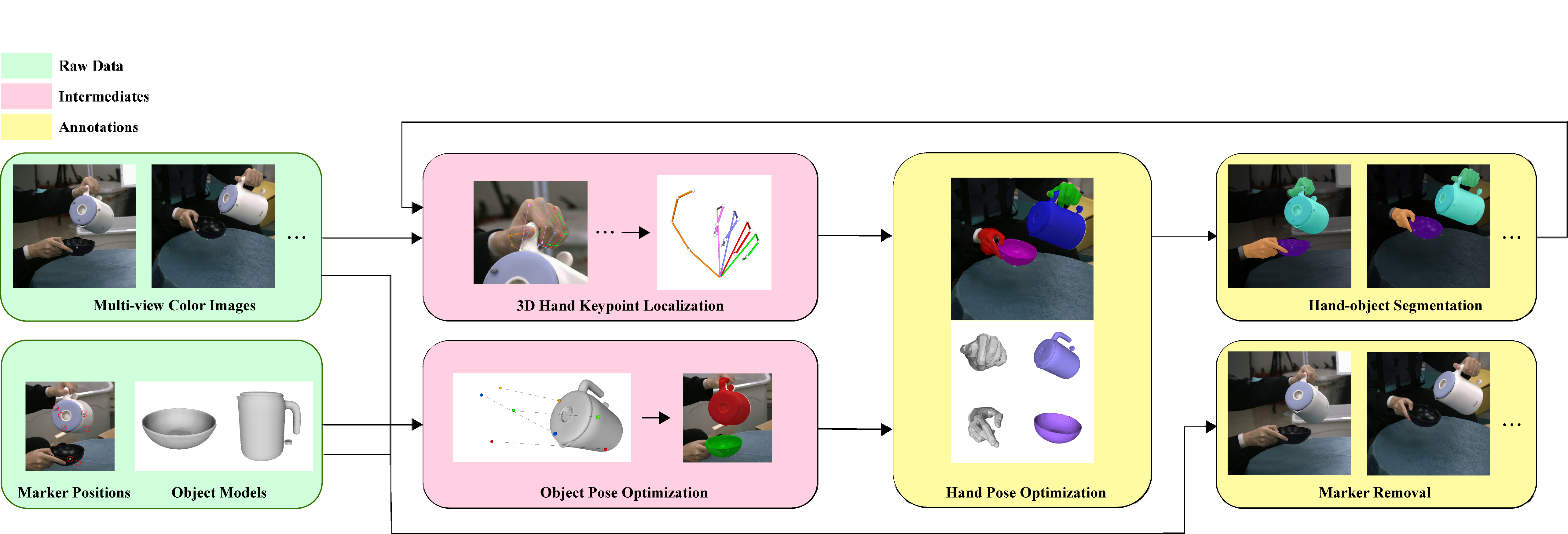}
   \vspace{-0.5cm}
   \caption{Automatic data annotating pipeline. The input consists of color frames from allocentric views, pre-scanned object models, and 3D positions of markers attached to object surfaces.  We first separately localize 3D hand keypoints and obtain object poses, and then conduct contact-aware optimization to recover MANO~\cite{mano} hand meshes. We finally segment hands and objects from images and automatically inpaint markers to acquire realistic object appearances.}
   \vspace{-0.5cm}
   \label{fig:data_annotating}
\end{figure*}


Figure \ref{fig:data_annotating} depicts our automatic data acquisition pipeline. Given color images, 3D marker positions, and object models as inputs, the pipeline successively performs object pose optimization, hand keypoint localization, hand pose optimization, hand-object segmentation, and marker removal.

\textbf{Object pose optimization.} Since our objects are rigid bodies, their motions are 6-dimensional and can be tracked by capturing marker positions using the mocap system. To accurately associate markers with the object, we draw inspiration from Mosh++~\cite{mahmood2019amass} and optimize marker-to-surface correspondence to encourage contact between markers and object surfaces while avoiding penetration. The object poses are obtained by combining marker positions relative to the object mesh and the captured marker motions.


\textbf{3D hand keypoint localization.} Leveraging multi-view sensing, we first estimate 2D hand keypoints from each camera view, respectively, and then fuse them into 3D. For 2D hand keypoint estimation, directly applying existing dual-hand approaches~\cite{h2o3d, IntagHand, jiang2023a2j} could fail when severe occlusion occurs or two hands are far apart. We thus follow ~\cite{openpose,mediapipe} and design a method to separately detect two hands using pre-trained YOLOv3~\cite{redmon2018yolov3} and then utilize MMPose~\cite{mmpose2020} to obtain single-hand 2D keypoints. To fuse multi-view 2D keypoints to 3D under severe occlusion, we use RANSAC~\cite{ransac} to filter out imprecise 2D keypoints and localize 3D keypoints by triangulation.


\textbf{Contact-aware hand pose optimization.}
We adopt MANO~\cite{mano} to formulate a 3D hand mesh as $\Theta_{h} = \{\theta, \beta, t\}$, where $\theta \in \mathbb{R}^{48}$, $\beta \in \mathbb{R}^{10}$, and $ t \in \mathbb{R}^{3}$ represent hand pose, hand shape, and wrist position, respectively. Given precomputed 3D hand keypoints and object meshes, we optimize MANO hand meshes by minimizing the following loss function:
\begin{align}
    \label{eq:eq1}
    \hat{\Theta}_{h} = &\underset{\Theta_{h}}{\arg \min } \left(\lambda_{2D} \mathcal{L}_{2D} + \lambda_{3D} \mathcal{L}_{3D} + \lambda_{angle} \mathcal{L}_{angle} +  \right.  \\[-0.5cm]
                       \nonumber
                       & \left. \lambda_{tc} \mathcal{L}_{tc} + \lambda_{p} \mathcal{L}_{p} + \lambda_{a} \mathcal{L}_{a}\right),
\end{align}
where $\mathcal{L}_{2D}$ and $\mathcal{L}_{3D}$ induce hand meshes to get close to the keypoints, $\mathcal{L}_{angle}$ provides hand pose priors, $\mathcal{L}_{tc}$ promotes temporal smoothness, $\mathcal{L}_{a}$ encourages contacts, and $\mathcal{L}_{p}$ prevents hand-object interpenetration.
Details of these loss terms are provided in the supplementary material.

\textbf{Hand-object segmentation.}
We obtain 2D hand-object masks by first rendering hand-object silhouettes from their 3D meshes. Since MANO meshes differ from real hands, we then apply the Segment Anything Model~\cite{segment_anything} to refine the masks, using sampled hand-object keypoints as prompts. To help detect hands in subsequent frames, we use Track-anything~\cite{track_anything} to track the hands in these frames and localize them.

\begin{figure}[h!]
  \centering
   \includegraphics[width=1.0\linewidth]{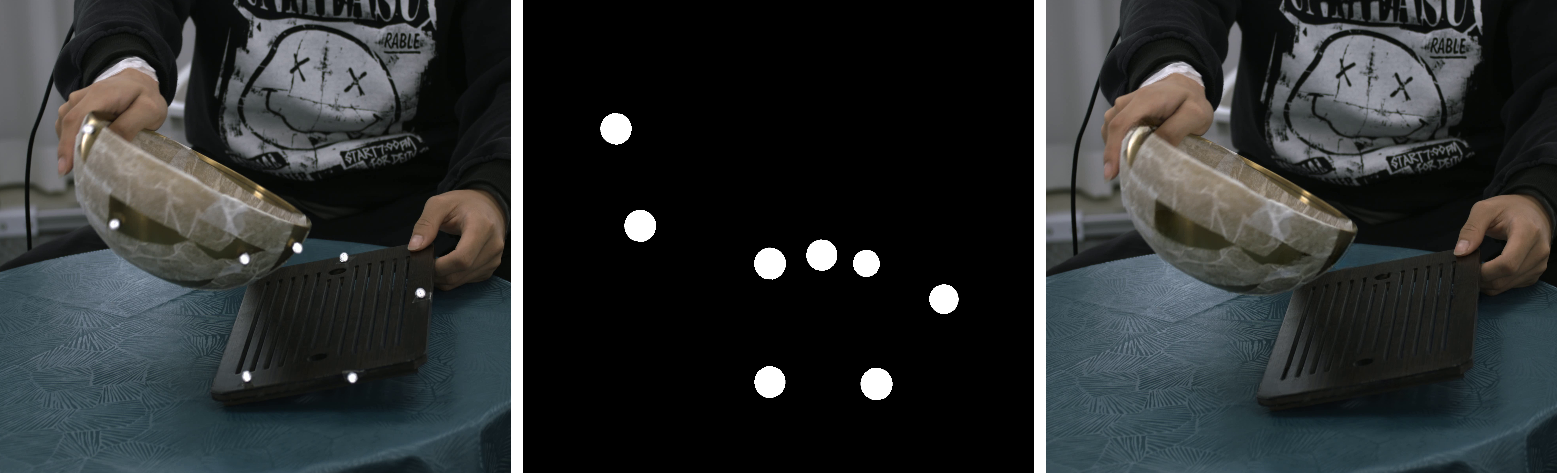}
   \caption{An example of marker removal. Three sub-figures respectively show the captured image patch, the automatically-computed marker mask, and the inpainted image patch.}
   \label{fig:marker_removal}
   \vspace{-0.4cm}
\end{figure}

\textbf{Marker removal.} Despite the fact that attaching markers to objects can accurately capture object motion, markers may increase the appearance gap between captured objects and those in the wild. One direct defect is that an object pose tracking network could simply track markers rather than learn to track the target object. To improve the object's appearance and make it more realistic, after acquiring hand and object poses, we automatically remove the markers from RGB images. First, we obtain marker positions in the world coordinate system from the mocap system, and place spheres with a radius two times the marker at those positions in a simulation environment. We then render a 2D mask of the spheres for each real camera view. Benefiting from the accurate marker position estimates, such 2D masks can fully cover the markers in the original RGB images. Finally, we mask the spheres in the original RGB images and use a pre-trained LAMA~\cite{lama} network to inpaint the sphere regions. Figure \ref{fig:marker_removal} shows a marker removal example.

\subsection{Dataset Statistics}
\label{sec:3.3}

\begin{figure}[h!]
  \centering
  \includegraphics[width=0.75\linewidth]{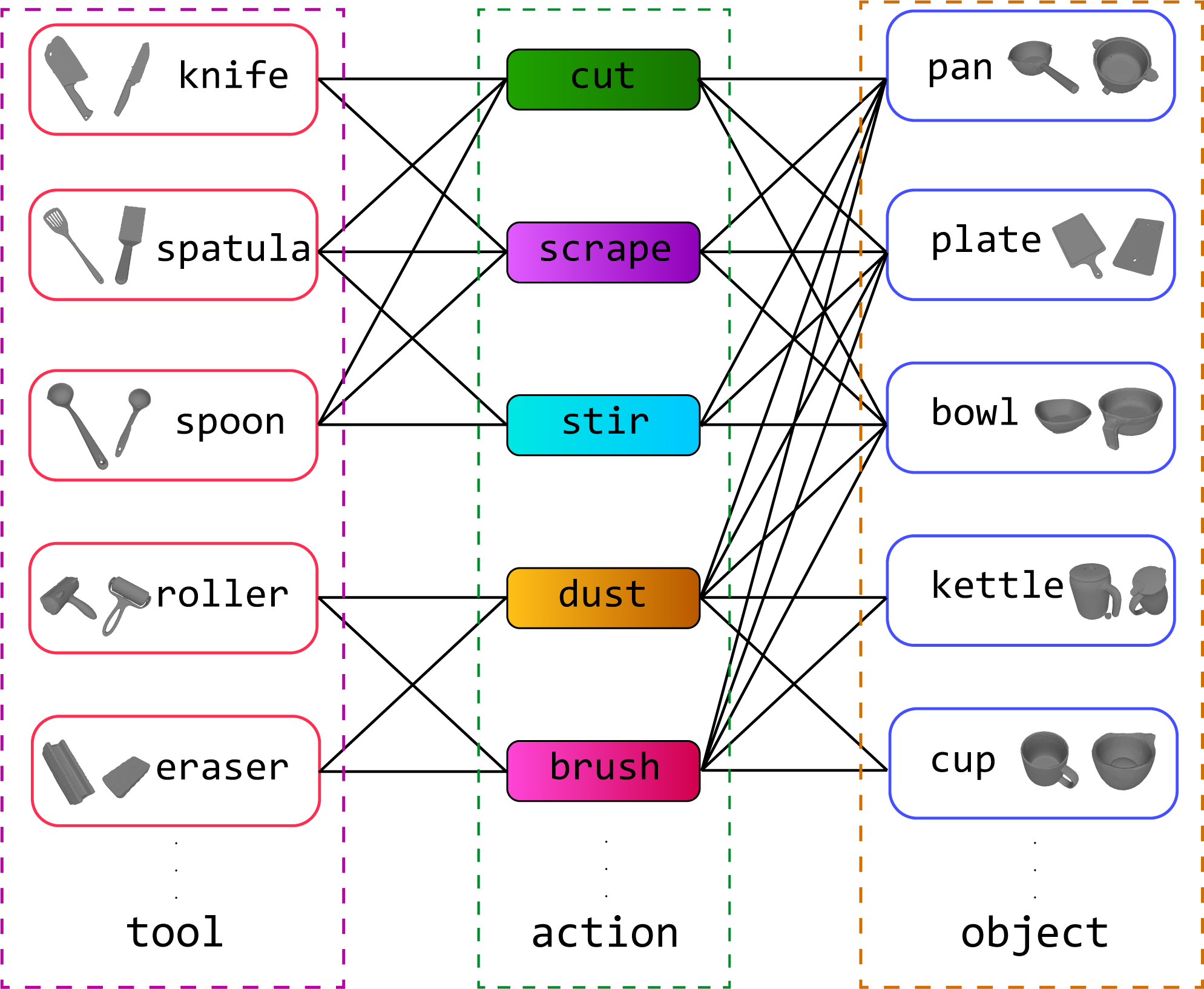}
  \vspace{-0.2cm}
   \caption{Examples of interaction triplets in \dataset. The left, middle, and right columns exemplify categories of tool, action, and target object, respectively. Triplets in our dataset are represented by connected paths from tools to target objects.}
   \vspace{-0.3cm}
   \label{fig:objects_and_actions}
\end{figure}

\dataset contains 2.5K motion sequences describing extensive daily tool-using behaviors covering 20 object categories, 196 fine-grained object 3D models, 14 participants, and 15 actions.
To support cross-category generalizable HOI studies, we explore the functional diversity of tools and target objects, with each tool and target object being utilized to perform up to 7 different kinds of actions. We formulate a behavior as using a \textbf{tool category} to perform an \textbf{action} on a \textbf{target object category} and divide our data into 131 $<$tool category, action label, target object category$>$ triplets. Some correlated triplets are exemplified in Figure \ref{fig:objects_and_actions}. To enable various studies, \dataset presents high-resolution color images from 12 allocentric views and RGBD images from one \textbf{egocentric-view}.

\section{Data Quality Evaluation}
\begin{figure}[h!]
    \centering
    \includegraphics[width=1.0\linewidth]{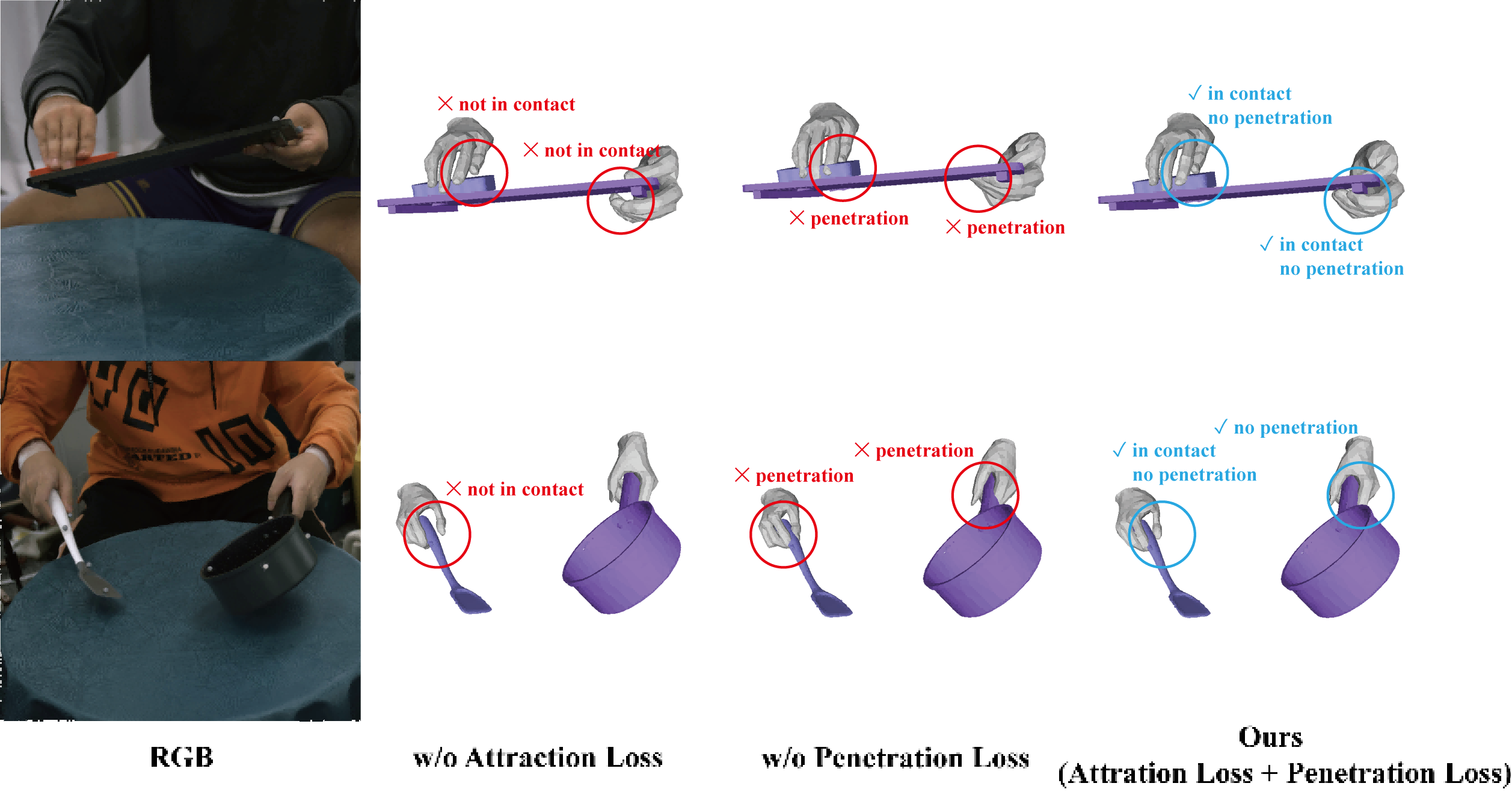}
    \vspace{-0.3cm}
    \caption{Qualitative contact optimization evaluation. From left to right: original RGB image, optimization without attraction loss, optimization without penetration loss, our method with attraction loss and penetration loss.}
    \vspace{-0.3cm}
   \label{fig:contact_loss_qualitative_evaluation}
\end{figure}

\textbf{Qualitative contact optimization evaluation.} Figure \ref{fig:contact_loss_qualitative_evaluation} depicts two frames of the TACO dataset. As shown in the figure, the attraction loss $\mathcal{L}_{a}$ promotes contact between the hands and the objects but may exacerbate penetration. The penetration loss $\mathcal{L}_{p}$ can mitigate penetration but does not address cases where the hands and the objects are not in contact. Our method combines both attraction loss and penetration loss, striking a balance between encouraging contact and preventing penetration.

\textbf{Quantitative hand pose evaluation.} To quantitatively examine the quality and value of our hand poses, we conduct a cross-dataset evaluation between \dataset and an existing high-quality HOI dataset DexYCB~\cite{chao2021dexycb}. We select a multi-stage method, CMR~\cite{CMR}, and a lightweight network, MobRecon~\cite{MobRecon}, both of which can recover occluded hand meshes from a single-view RGB image captured in HOI scenarios, and then train and test their networks under different data combinations.
For DexYCB, we follow its default train/test split (S0).
For \dataset, we randomly select 60 motion sequences for training and 40 for testing and crop the right hand from each image to align with the single-hand HO3D. Following existing hand mesh recovery methods~\cite{CMR,MobRecon,park2022handoccnet,xu2023h2onet}, we select Procrustes-Aligned Mean Per Joint Position Error (PA-MPJPE) and Mean Per Joint Position Error (MPJPE) as evaluation metrics. More details about the metrics can be found in the paper~\cite{MobRecon}. Table \ref{tab:cross_dataset_validation} shows the performance of the two methods under different settings. Due to the large gap in hand-object motion between the two datasets, the performances for the two methods would both drop rapidly when the training and the test sets are from different datasets. Nevertheless, combining training data from \dataset and DexYCB achieves significant performance gains on both datasets for both the two methods, which demonstrates the accuracy of our hand pose annotations and indicates that \dataset is a beneficial data complement for 3D hand pose estimation.

\begin{table}[t]
\centering
\footnotesize
\addtolength{\tabcolsep}{-3pt}
{
\begin{tabular}{|c|c|c|c|c|}
\hline  
Method & \backslashbox{Test}{Train} & \dataset & DexYCB & \makecell{\dataset+\\DexYCB} \\
\hline
\multirow{2}{*}{CMR} & \dataset & 10.73 / 30.87 & 13.53 / 39.54  & \textbf{10.25} / \textbf{28.36} \\
\cline{2-5}
& DexYCB & 19.36 / 76.20 & 6.51 / 13.92 & \textbf{6.44} / \textbf{13.62} \\
\hline
\multirow{2}{*}{MobRecon} & \dataset & 9.03 / 20.76 & 12.95 / 34.32 & \textbf{8.69} / \textbf{20.14} \\
\cline{2-5}
& DexYCB & 15.17 / 44.73 & 6.51 / 13.61 & \textbf{6.32} / \textbf{12.94} \\
\hline
\end{tabular}
}
\vspace{-0.3cm}
\caption{Cross-dataset evaluation with DexYCB~\cite{chao2021dexycb} on 3D hand pose estimation. Results are in PA-MPJPE (mm, lower is better) and MPJPE (mm, lower is better), respectively.}
\vspace{-0.7cm}
\label{tab:cross_dataset_validation}
\end{table}

\begin{figure}[h!]
  \centering
   \includegraphics[width=1.0\linewidth]{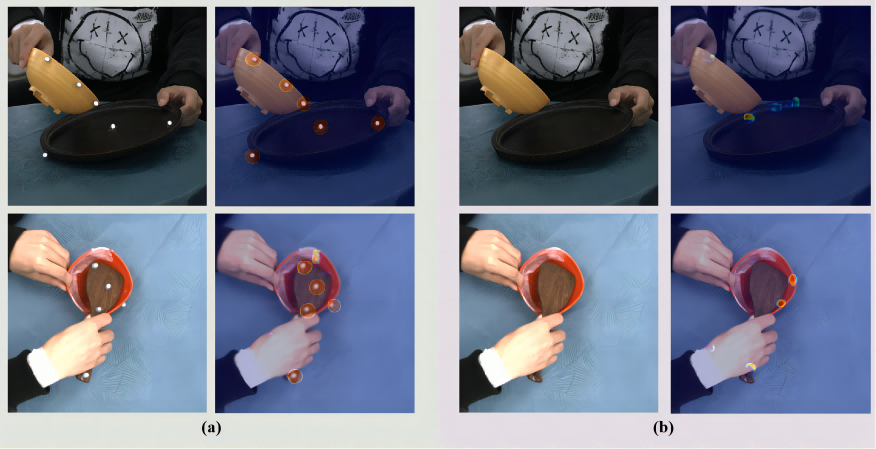}
   \vspace{-0.3cm}
   \caption{Marker removal quality evaluation. We train and test U-Net~\cite{U-Net} on raw captured images (a) and marker-removed images (b), respectively. In each experiment, the left and right columns show the input image patches and the predicted segmentation heatmaps, respectively.}
   \vspace{-0.5cm}
   \label{fig:marker_removal_quality_check}
\end{figure}

\textbf{Marker removal quality evaluation.} We conduct a novel experiment to examine the reality of our marker-removed images. The key idea is that a network can easily segment markers from the image while being hard to figure out the same regions when markers are replaced by the original object surface appearances. We thus use the originally captured image patches and the marker-removed ones to train a 2D marker segmentation network, respectively, and then apply the two networks to novel image patches with the same image processing setting as training. In practice, spheres larger than markers (as shown in Figure \ref{fig:marker_removal}(b)) are required to be segmented, since the marker removal process exactly inpaints these regions. We select U-Net~\cite{U-Net} as the network backbone and train the two networks with an L2-loss measuring differences between mask predictions and ground truths. When training and testing on raw image patches, U-Net achieves 63.8\% mean Intersection over Union (mIoU), while its performance sharply drops to 11.1\% mIoU after applying marker removal, indicating the reality and naturalness of the inpainted image regions. Figure \ref{fig:marker_removal_quality_check} further compares heatmaps of network predictions under the two settings, exhibiting ambiguities of marker segmentation posed by realistic inpainted images. More details are provided in the supplementary material.

\section{Experiments}
In this section, we first introduce our data split (Section \ref{sec:data_split}) supporting different research purposes about generalization. We then present three benchmarks on \dataset: compositional action recognition (Section \ref{sec:compositional_action_regocnition}), generalizable hand-object motion forecasting (Section \ref{sec:motion_forecasting}), and cooperative grasp synthesis (Section \ref{sec:grasp_synthesis}). Details of evaluation metrics are provided in supplementary materials.

\subsection{Data Split}
\label{sec:data_split}

With diverse object geometries and interaction triplets, \dataset focuses on supporting generalizable studies that apply to unseen object geometries or novel interaction triplets. Hence, we carefully split our test data into four subsets with different generalization purposes:

\begin{itemize}

\item Test set 1 (S1): \textit{No generalization.} Tool geometries and interaction triplets are both involved in the training set.
\item Test set 2 (S2): \textit{Geometry-level generalization.} The tool geometry is novel, while the interaction triplet appears in the training set.
\item Test set 3 (S3): \textit{Interaction-level generalization.} The interaction triplet is novel, while the tool categories and geometries are included in the training set.
\item Test set 4 (S4): \textit{Compound generalization.} The tool category is novel, leading to unseen geometries and triplets.

\end{itemize}

The training set and the four test sets follow a data ratio of 4:1:1:1.5:2.5.

\subsection{Compositional Action Recognition}
\label{sec:compositional_action_regocnition}

Humans can recognize the action disentangled from their context and appearance biases of the objects, which is referred to as compositionality. To test a system’s compositional generalization capability, with a diverse array of tool-action-object triplets in TACO, we benchmark compositional action recognition aiming at identifying actions in HOI videos. In contrast to traditional action recognition tasks~\cite{soomro2012ucf101,shahroudy2016ntu,goyal2017something,epic-kitchens}, our focus is on scenarios where the tool-action-object triplets in the test set are unseen during training, including novel tool types or geometries. This poses additional challenges for evaluating model generalization and achieving human-like perception.

\textbf{Problem formulation:}
 In a bimanual HOI scenario, our goal is to identify action labels from first-person RGB video frames, with optional bounding boxes for hands and objects computed based on their poses.

\textbf{Evaluation metrics:} Following~\cite{yang2023aim,radevski2021revisiting}, we use Top-1 Accuracy and Top-5 Accuracy to evaluate the efficacy of action recognition.

\textbf{Baselines, results, and analysis:} We compare two baselines to evaluate the compositional generalization capability: AIM~\cite{yang2023aim}, an adaptation of pre-trained image transformer models, representing state-of-the-art traditional action recognition approaches; and CACNF~\cite{radevski2021revisiting}, which specializes in compositional and few-shot action recognition. Table \ref{tab:action_recognition} shows the results. A significant decrease in accuracy on the most difficult S4 set, compared to high accuracy on the non-generalizable S1 set, underscores compositional generalization challenges. Notably, the S2 set, with novel tool geometries within tool categories, performed similarly to S1, while the S3 set, featuring novel tool-action-object triplets, showed a significant accuracy drop, revealing that while action recognition models inherently possess basic generalizability to geometric changes, they struggle with novel triplets due to the unfamiliarity of interaction contexts. When comparing among baselines, CACNF outperformed AIM in challenging sets S3 and S4, benefiting from its bounding box-based approach that effectively disentangles actions from object and tool geometries, highlighting the advantages of a compositional approach in generalizable cases. All these results underscore the significant impact of compositional generalization on action recognition.  

\begin{table}[t]
\centering
\small
\addtolength{\tabcolsep}{-3pt}
{
\begin{tabular}{|c|c|c|c|}
\hline  
Test Set & Method & Top-1 (\%, $\uparrow$) & Top-5 (\%, $\uparrow$)  \\
\hline
\multirow{2}{*}{S1} & AIM~\cite{yang2023aim} & 83.08 & 98.85 \\
& CACNF~\cite{radevski2021revisiting} & \textbf{86.15} & \textbf{99.62}\\

\hline
\multirow{2}{*}{S2} & AIM~\cite{yang2023aim} & \textbf{82.81} & \textbf{98.83} \\
& CACNF~\cite{radevski2021revisiting} & 77.34 & 96.88\\

\hline
\multirow{2}{*}{S3} & AIM~\cite{yang2023aim} & 53.65 & 81.25 \\
& CACNF~\cite{radevski2021revisiting} & \textbf{63.02} &\textbf{92.97} \\

\hline
\multirow{2}{*}{S4} & AIM~\cite{yang2023aim} & 39.33 & 73.67 \\
& CACNF~\cite{radevski2021revisiting} & \textbf{44.00} & \textbf{80.50} \\

\hline
\end{tabular}
}
\vspace{-0.2cm}
\caption{Results on compositional action recognition. Methods are examined via Top-1 and Top-5 accuracy.}
\vspace{-0.5cm}
\label{tab:action_recognition}
\end{table}

\subsection{Generalizable Hand-object Motion Forecasting}
\label{sec:motion_forecasting}

With abundant 3D hand-object mesh annotations provided by \dataset, we benchmark generalizable interaction forecasting aiming to predict the following hand-object motions from seen short motion clips. Unlike the prediction of low-frequency human motions~\cite{peng2022somoformer,mao2022contact,peng2023trajectory}, we observe that hands commonly achieve a complete manipulation behavior (e.g. pouring all the water out of a bowl) in a very short time, making forecasting both interesting and challenging.

\textbf{Problem formulation:} In a bimanual HOI scenario, given object point clouds and poses of two hands, the tool, and the target object in consecutive $N$ frames, our goal is to forecast their poses in subsequent $M$ frames. We select $N$=10 and $M$=10 in our dataset.

\textbf{Evaluation metrics:} Following human-object forecasting evaluations~\cite{CAHMP,adeli2021tripod,HO-GCN,InterDiff}, we use Mean Per Joint Position Error $J_e$ to evaluate predicted left and right-hand skeletons, and leverage translation error $T_e$ and rotation error $R_e$ to measure object positions and orientations, respectively.

\textbf{Baselines, results, and analysis:} Due to the lack of hand-object interaction forecasting solutions, we transfer methodologies from human-object motion forecasting as baselines. To cover diverse method designs and possibilities, we select two generative models~\cite{MDM,InterDiff} and two predictive models~\cite{CAHMP,InterDiff} for comparison. Table \ref{tab:motion_forecasting} compares their performance under different generalization settings. We observed that the tool and the hand holding it play dominant roles during manipulation, making their motion forecasting more challenging compared to others. The generative models, in comparison to predictive ones, yield significantly larger errors for both hands and the tool. One possible reason is that their motion patterns are fast and complex, making the modeling of motion distribution challenging. Enhancing the accuracy of generative methods in hand-object manipulation scenarios is an interesting future direction. When comparing method performances between a non-generalizable set (S1) and sets involving generalization (S2-4), we note that methods consistently exhibit significant performance declines with the right hand and the tool, regardless of whether applied to novel tool geometries (S2, S4) or interaction triplets (S3, S4), meanwhile achieving close results on the others. This trend indicates substantial challenges in generalizing dominant interaction entities.

\begin{table}[t]
\centering
\small
\addtolength{\tabcolsep}{-3pt}
{
\begin{tabular}{|c|c|c|c|c|}
\hline  
Test & \multirow{2}{*}{Method} & $J_e$ (mm, $\downarrow$) & $T_e$ (mm, $\downarrow$) & $R_e$ ($^\circ$, $\downarrow$) \\
\cline{3-5}
Set & & Right / Left & Tool / Target & Tool / Target \\
\hline
\multirow{4}{*}{S1} & InterVAE~\cite{InterDiff} & 54.9 / 48.9 & 55.0 / 15.8 & 66.94 / 6.63 \\
& MDM~\cite{MDM} & 61.2 / 53.7 & 49.4 / 14.9 & 52.81 / 5.85 \\
& InterRNN~\cite{InterDiff} & 29.3 / \textbf{20.4} & 25.8 / \textbf{11.6} & \textbf{10.08} / 5.22 \\
& CAHMP~\cite{CAHMP} & \textbf{28.8} / 22.1 & \textbf{23.9} / 12.3 & 10.24 / \textbf{4.95} \\
\hline
\multirow{4}{*}{S2} & InterVAE~\cite{InterDiff} & 58.8 / 50.8 & 54.3 / 12.2 & 88.19 / 5.24 \\
& MDM~\cite{MDM} & 65.8 / 55.8 & 48.0 / 10.1 & 75.78 / 3.59 \\
& InterRNN~\cite{InterDiff} & 35.2 / 23.1 & 31.0 / 8.9 & 11.09 / 4.03 \\
& CAHMP~\cite{CAHMP} & \textbf{31.4} / \textbf{22.6} & \textbf{24.7} / \textbf{8.6} & \textbf{10.65} / \textbf{3.32} \\
\hline
\multirow{4}{*}{S3} & InterVAE~\cite{InterDiff} & 56.5 / 50.1 & 56.7 / 12.7 & 68.03 / 5.69 \\
& MDM~\cite{MDM} & 66.4 / 58.0 & 46.2 / 11.5 & 61.90 / 4.14 \\
& InterRNN~\cite{InterDiff} & 32.5 / 22.0 & 28.2 / 9.7 & 11.42 / 4.32 \\
& CAHMP~\cite{CAHMP} & \textbf{29.3} / \textbf{21.6} & \textbf{24.4} / \textbf{9.4} & \textbf{11.18} / \textbf{3.80} \\
\hline
\multirow{4}{*}{S4} & InterVAE~\cite{InterDiff} & 63.9 / 54.2 & 63.0 / 14.5 & 71.38 / 5.01 \\
& MDM~\cite{MDM} & 70.5 / 57.9 & 50.2 / 12.2 & 70.25 / 3.55 \\
& InterRNN~\cite{InterDiff} & 36.6 / 22.3 & 30.9 / 10.6 & 11.92 / 3.86 \\
& CAHMP~\cite{CAHMP} & \textbf{32.9} / \textbf{22.2} & \textbf{26.6} / \textbf{10.8} & \textbf{11.73} / \textbf{3.24} \\
\hline
\end{tabular}
}
\vspace{-0.2cm}
\caption{Results on generalizable interaction forecasting. We respectively examine predictions of the right hand, the left hand, the tool, and the target object.}
\vspace{-0.5cm}
\label{tab:motion_forecasting}
\end{table}

\subsection{Cooperative Grasp Synthesis}
\label{sec:grasp_synthesis}

Generating human-like hand grasps benefits various applications in VR/AR and dexterous manipulation. Existing grasp synthesis approaches~\cite{grasp_tta,HALO,TOCH,ContactGen} mainly focus on creating stable grasps on static objects without considering interactive purposes. Taking advantage of extensive interaction behaviors from \dataset, we benchmark a novel task that aims to synthesize realistic and physically plausible hand grasps in HOI scenarios. Besides achieving stable grasping for the object, the method must comprehend other interactive objects and human hands to generate cooperative motions and avoid conflicts. We evaluate this task on our four test sets, each targeting different aspects of generalization.

\textbf{Problem formulation:} Consider using the right hand to manipulate a tool and the left hand to operate a target object. During manipulation, given the meshes of the two objects and the left hand, the goal is to generate a right-hand mesh grasping the tool in a manner conducive to the interaction.

\textbf{Evaluation metrics:} Following static grasp synthesis methods~\cite{grasping_field,grasp_tta,ContactGen}, we use interpenetration volume (\textit{Pen. V}) and contact ratio (\textit{Con. R}) to measure the physical plausibility of generated hands. To ensure the hand is appropriately interactive, we introduce a new metric named collision ratio (\textit{Col. R}), which detects collisions between generated hands and the interaction environment. Nevertheless, to quantitatively examine whether the grasps are realistic, we design the FID score (\textit{FID}) to measure distances between distributions of human grasps and synthesized ones.

\begin{figure}[h!]
    \centering
    \includegraphics[width=1.0\linewidth]{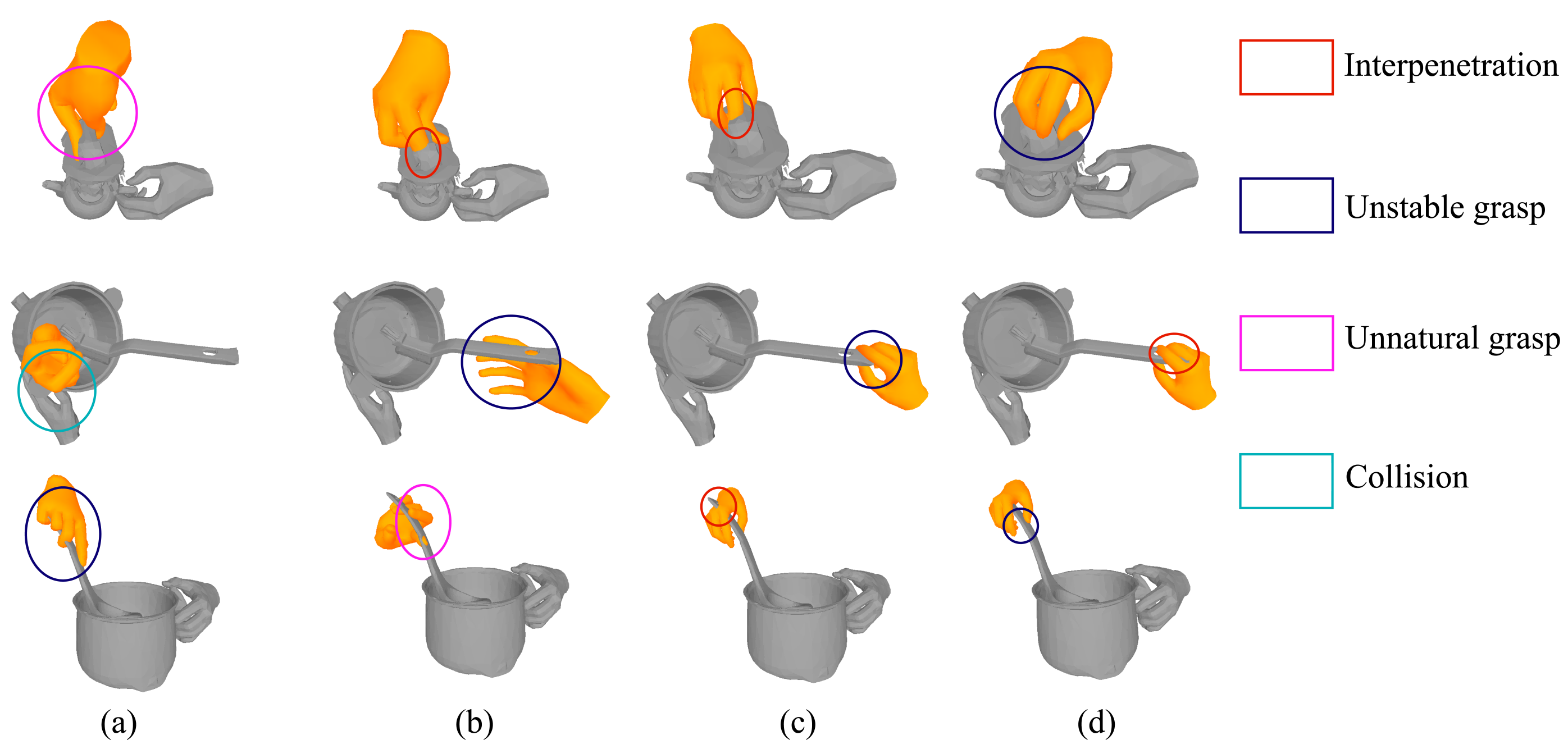}
    \vspace{-0.3cm}
    \caption{Failure cases of cooperative grasp synthesis. (a) and (b) are generated by ContactGen~\cite{ContactGen}, while (c) and (d) are from HALO-VAE~\cite{HALO}.}
    \vspace{-0.5cm}
    \label{fig:grasp_synthesis}
\end{figure}

\textbf{Baselines, results, and analysis:} We select two recent CVAE-based approaches ContactGen~\cite{ContactGen} and HALO-VAE~\cite{HALO}, and modify their network structures to incorporate the left hand and the target object as additional VAE conditions. To assess the significance of interaction environments, we also evaluate HALO-VAE$^-$, which omits the use of environment inputs and directly generates from the tool mesh. Due to the strong correlation between hand grasps and tool geometries, we test these approaches with familiar (S1, S3) and unseen tool geometries (S2, S4). Table \ref{tab:grasp_synthesis} summarizes quantitative evaluations of the two settings. Notably, HALO-VAE$^-$ shows a significant decrease in performance on contact and collision ratios compared to HALO-VAE, underscoring the importance of perceiving and modeling interaction environments. When applied to novel object geometries and categories, all three approaches exhibit higher collision ratios and FID scores, and most of them show a marked decline in their ability to contact tools precisely and prevent penetrations. Figure \ref{fig:grasp_synthesis} exemplifies failure cases of synthesized grasps, indicating a large room for improving physical feasibility and naturalness on complex and slender geometries.

\begin{table}[t]
\centering
\small
\addtolength{\tabcolsep}{-3pt}
{
\begin{tabular}{|c|c|c|c|c|c|}
\hline  
\multirow{2}{*}{Test Set} & \multirow{2}{*}{Method} & \textit{Pen. V} & \textit{Con. R} & \textit{Col. R} & \textit{FID} \\
 & & (cm$^3$, $\downarrow$) & ($\%$, $\uparrow$) & ($\%$, $\downarrow$) & ($10^{-2},\downarrow$) \\
\hline
\multirow{3}{*}{S1 $\bigcup$ S3} & ContactGen~\cite{ContactGen} & 0.27 & 80.12 & 10.40 & 12.18 \\
 & HALO-VAE$^-$ & 0.31 & 76.27 & 4.19 & 4.89 \\
 & HALO-VAE~\cite{HALO} & \textbf{0.26} & \textbf{82.86} & \textbf{2.39} & \textbf{2.02} \\
\hline
\multirow{3}{*}{S2 $\bigcup$ S4} & ContactGen~\cite{ContactGen} & \textbf{0.26} & 74.98 & 17.56 & 22.53 \\
 & HALO-VAE$^-$ & 0.29 & 71.69 & 14.25 & 5.09 \\
 & HALO-VAE~\cite{HALO} & 0.32 & \textbf{83.52} & \textbf{11.11} & \textbf{2.84} \\
\hline
\end{tabular}
}
\vspace{-0.2cm}
\caption{Results on cooperative grasp synthesis. We measure the physical plausibility of generated grasps by interpenetration volume and contact ratio, meanwhile using collision ratio to examine interaction faithfulness and FID score to assess reality.}
\vspace{-0.5cm}
\label{tab:grasp_synthesis}
\end{table}

\section{Limitations and Conclusion}
We present \dataset, the first large-scale, real-world 4D bimanual hand-object manipulation dataset. It encompasses a wide range of tool-action-object compositions and object geometries. \dataset contains a total of 2.5K motion sequences and 5.2M video frames, captured from 12 third-person views and one egocentric view. We contribute an automatic data acquisition pipeline that accurately recovers hand-object meshes and segmentation, along with realistic hand-object appearances. Leveraging \dataset's diverse data, we benchmark compositional action recognition, generalizable hand-object motion forecasting, and cooperative grasp synthesis. These benchmarks reveal new insights, challenges, and opportunities in the field of generalizable hand-object interaction studies.

There are three major limitations in \dataset. Firstly, \dataset currently does not cover articulated objects. Secondly, while \dataset offers an extensive exploration of object geometries and HOI behaviors, it lacks scene diversities that are also crucial for understanding human manipulations. Thirdly, our solution to marker removal is an application of generative models, hence the original object appearances cannot be recovered perfectly.

{
    \small
    \bibliographystyle{ieeenat_fullname}
    \bibliography{ref}
}

\clearpage
\Large \textbf{Appendix}
\normalsize

\section{Interaction Field Estimation}
\label{sec:interaction_field_estimation}

With accurate object models and MANO\cite{mano} meshes captured in our dataset, we benchmark estimating interaction fields of hands and objects from color images. Apart from recovering hand-object interaction fields in existing work\cite{fan2023arctic}, our task also involves estimating those between tools and target objects.

\textbf{Problem formulation:} Representing the left hand, right hand, tool, and target object with $l$, $r$, $t$, and $o$, the task is to estimate six interaction fields between hands and objects ($F^{r \rightarrow t}$, $F^{t \rightarrow r}$, $F^{l \rightarrow o}$, $F^{o \rightarrow l}$, $F^{t \rightarrow o}$, $F^{o \rightarrow t}$) from a given RGB frame, where field $F^{a \rightarrow b}$ is defined as the distance to the nearest vertex in mesh $b$ for all vertices in mesh $a$.

\textbf{Evaluation metrics:} Following \cite{fan2023arctic}, we use the Mean Distance Error to evaluate the precision of predicted interaction fields, and the Acceleration Error to measure the smoothness of those estimates.

\textbf{Baselines, results and analysis:} We set up two baseline methods based on InterField-SF\cite{fan2023arctic}. The first one (InterField-SF separated) takes an image and two meshes as input (e.g. the right hand and the tool) and estimates the two fields between them. The second one (InterField-SF concatenated) incorporates the image with meshes of both hands, the tool, and the target object, and predicts six interaction fields altogether. Table \ref{tab:interaction_field_estimation} shows the quantitative results of the two methods. The primary distinctions among test sets lie in the selection of tools and actions, with a relatively weaker correlation to target objects. The results suggest that these variations exert a more pronounced influence on RT and TR. Within the fields of RT and TR, the method performances on S1 significantly surpass those on S3, while the latter outperforms both S2 and S4. This indicates that the methods face challenges in generalizing to unseen geometries. The incorporation of seen geometries with unseen actions (S3) also introduces additional complexities. These challenges in generalization to both unseen geometries and actions underscore the need for further exploration and refinement of the proposed methods.

\begin{table*}[t]
\centering
\small
\addtolength{\tabcolsep}{0pt}
{
\begin{tabular}{|c|c|c|c|c|c|c|c|c|c|c|c|c|c|}
\hline  
Test Set & Method & \multicolumn{6}{c|}{Mean Distance Error (mm, $\downarrow$)} & \multicolumn{6}{c|}{Acceleration Error ($m/s^2$, $\downarrow$)}  \\
\cline{3-14}
& & RT & TR & LO & OL & TO & OT & RT & TR & LO & OL & TO & OT \\
\hline
\multirow{2}{*}{S1} & InterField-SF (separated) & 8.4 & 8.7 & 10.4 & 21.5 & \textbf{11.6} & \textbf{15.7} & \textbf{8.7} & \textbf{8.3} & \textbf{10.3} & 12.3 & \textbf{13.7} & \textbf{16.3}  \\
& InterField-SF (concatenated) & \textbf{8.1} & \textbf{8.5} & \textbf{10.1} & \textbf{21.3} & 12.7 & 17.3 & 9.0 & 8.6 & 10.4 & \textbf{11.9} & 14.6 & 17.7 \\
\hline
\multirow{2}{*}{S2} & InterField-SF (separated) & \textbf{14.9} & 32.5 & 14.7 & \textbf{18.9} & 15.6 & \textbf{12.5} &  \textbf{10.5} & 11.4 & \textbf{12.0} & \textbf{10.7} & \textbf{13.8} & \textbf{11.6} \\
& InterField-SF (concatenated) & 15.1 & \textbf{31.5} & \textbf{14.6} & 19.1 & \textbf{15.2} & 12.8 & 10.7 & \textbf{11.3} & 12.2 & 10.9 & 14.4 & 12.6  \\
\hline
\multirow{2}{*}{S3} & InterField-SF (separated) & 13.6 & 17.2 & \textbf{15.1} & \textbf{26.6 }& \textbf{13.0} & \textbf{14.0} & \textbf{9.8} & 9.1 & 11.2 & \textbf{10.8} & \textbf{13.0} & \textbf{13.4} \\
& InterField-SF (concatenated) & \textbf{13.4} & \textbf{16.9} & 15.2 & 27.1 & 14.3 & 14.2 & 9.9 & \textbf{9.0} & \textbf{11.1} & 11.1 & 13.9 & 14.3  \\
\hline
\multirow{2}{*}{S4} & InterField-SF (separated) & \textbf{13.9} & \textbf{34.2} & \textbf{12.5} & \textbf{19.0} & \textbf{15.0} & \textbf{12.5} & 12.6 & \textbf{9.6} & \textbf{10.7} & \textbf{10.3} & \textbf{14.5} & \textbf{13.3} \\
& InterField-SF (concatenated) & 14.0 & 35.4 & 12.6 & 19.3 & 15.2 & 13.0 & \textbf{10.6} & 10.0 & 11.2 & \textbf{10.3} & 15.4 & 14.5  \\
\hline
\end{tabular}
}
\vspace{-0.3cm}
\caption{Results on interaction field estimation\cite{fan2023arctic}, where R denotes right hand, T denotes tool, L denotes left hand and O denotes target object (e.g. RT means right-hand-to-tool). Methods are examined via Mean Distance Error and Acceleration Error.}
\vspace{-0.5cm}
\label{tab:interaction_field_estimation}
\end{table*}

\section{Data Capturing Details}

\subsection{Camera Calibration}

\textbf{Camera intrinsic calibration.} We use a traditional method that places a checkerboard in the camera view with known scales of grids and estimates the camera intrinsic matrix and distortion using OpenCV functions.

\begin{figure}[h!]
    \centering
    \includegraphics[width=0.8\linewidth]{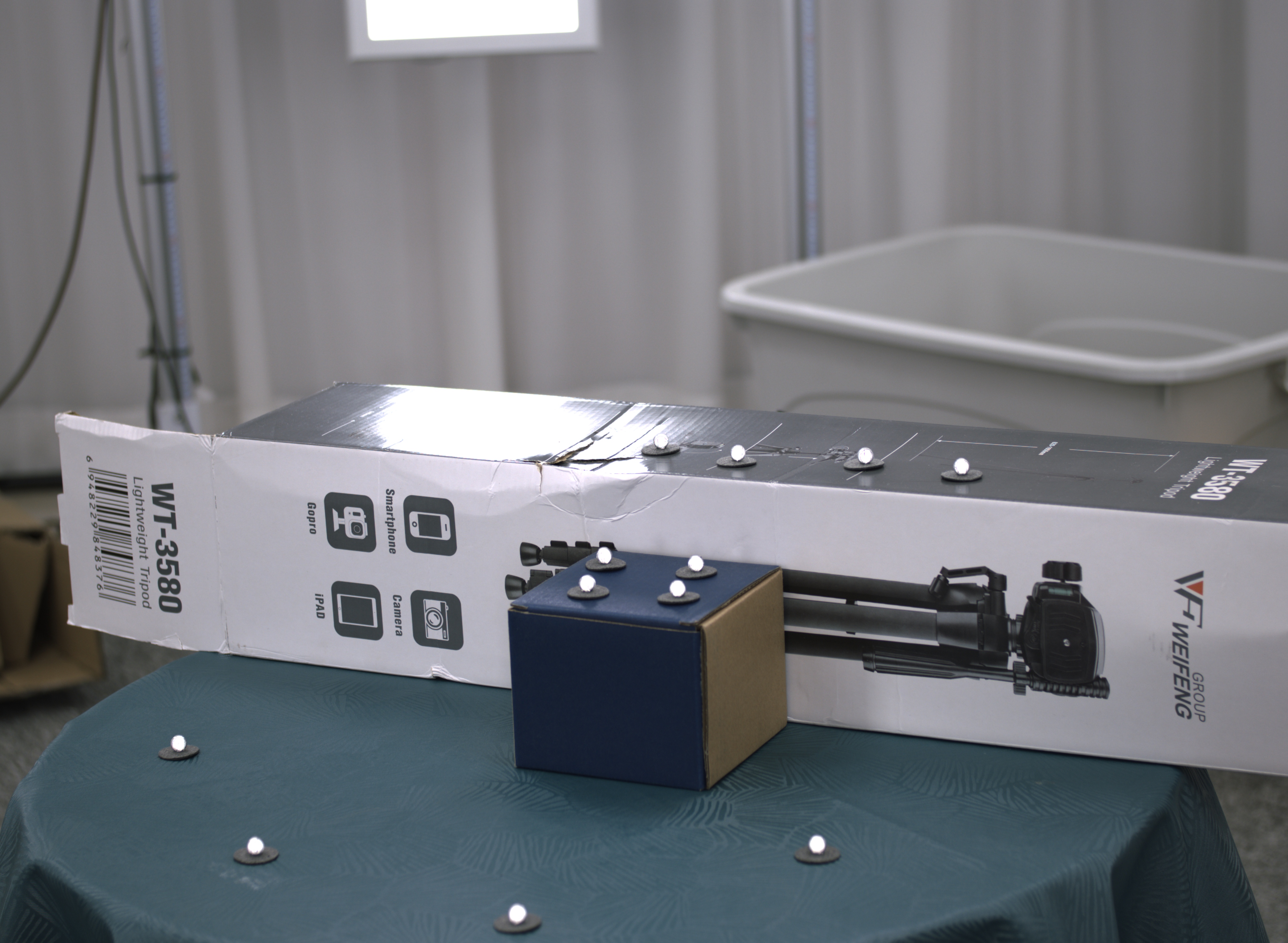}
    \vspace{-0.3cm}
    \caption{Calibrating the camera extrinsic.}
    \vspace{-0.5cm}
    \label{fig:camera_calib}
\end{figure}

\textbf{Camera extrinsic calibration.} After acquiring the camera intrinsic, we perform a semi-automatic process for calibrating the camera extrinsic before data capturing. As shown in Figure \ref{fig:camera_calib}, we first place 12 markers in the scene. Benefiting from our mocap system, we can obtain accurate marker positions in the world coordinate system with errors less than 1mm. We then manually annotate the pixel coordinate of each marker in the color image, and compute the optimal camera extrinsic minimizing re-projection error of markers. We solve this Perspective-n-Point (PnP) problem with OpenCV algorithms.

\subsection{Time Synchronization}

We provide time-synchronized data from the different
sensor modalities. 
Our 12 industrial FLIR cameras receive signals from the same signal generator through audio cables.
To synchronize industrial cameras with our mocap system and Realsense L515 camera, we record UTC timestamps for each frame captured by different cameras and perform nearest-neighbor matching among timestamps. 
The maximal time difference between matched signals is 17ms.

\section{Data Annotating Details}

\subsection{Details on Object Pose Optimization}

We attach four markers with a radius of 4mm to the surface of each object and obtain the object pose by capturing marker positions by the optical mocap system. To reuse the markers and optimization results, we mark a target position on the object surface for each marker and attach markers to these fixed positions before data collection. For each object, we formulate the attached four markers as a rigid body $B$ and optimize the relative 6D pose $T=[R,t]$ from the 3D object model to $B$, where $R \in SO(3)$ denotes 3D rotation, and $t \in \mathbb{R}^{3}$ indicates 3D translation. Since markers actually contact the object surface without interpenetration, we first design contact loss $L_c(q,P)$ and penetration loss $L_p(q,P)$ as:

\vspace{-0.5cm}
\begin{equation}
\begin{split}
& L_c (q, P) = \Vert q - p_{i^*} \Vert_2, \\
& L_p (q, P) = \max(- \vec{n}_{i^*}^T (q - p_{i^*}), 0),
\end{split}
\end{equation}

where $q \in \mathbb{R}^3$ is a query point, $P = \{p_i \in \mathbb{R}^3\}_{i=1}^{|P|}$ is a point cloud, $i^*=\mathop{\arg\min}\limits_{1 \leq i \leq |P|} \Vert q - p_i \Vert_2$ denotes the index of the closest point in $P$ to $q$, and $\vec{n}_i$ denotes the normal of the point $p_i$. We then incorporate the two loss functions and compute the optimal relative pose $T^*$ via the following function:

\vspace{-0.5cm}
\begin{equation}
\begin{split}
T^* = \mathop{\arg\max}\limits_{R,t} \sum_{k=1}^{K} & (  L_c(R q_k + t, P) + L_p(R q_k + t, P) ) \\
 & [L_c (R q_k + t, P) < \alpha],
\end{split}
\end{equation}

where $K$ is the number of markers, $q_k \in \mathbb{R}^3$ is the marker position in the coordinate system of $B$, $P$ is the vertices of the object model, and $\alpha$=1cm is a threshold selecting markers near the object. Given a manual initialization of $T$, we use the Adam optimizer to find $T^*$ with learning rate 1e-4. In practice, we attach 10 additional markers to the object surface ($K$=14) to improve the robustness of the optimization, meanwhile using only four of them to track the object during data capturing.

\subsection{Details on 3D Hand Keypoint Localization}

For the initial frame of the entire sequence, we employ a pre-trained YOLOv3\cite{redmon2018yolov3} to obtain the bounding boxes for both left and right hands. 
For subsequent frames, we leverage the Track-Anything Model\cite{track_anything} along with the optimized hand pose from the preceding frame to generate masks for both hands and compute bounding boxes based on these masks.
Then, we crop out sub-images containing only one hand according to these bounding boxes. 
The resulting sub-images undergo processing via the single-hand pose estimation model MMPose\cite{mmpose2020} to determine 2D keypoint positions $K_{2D_{c}}[i]$ for each hand in each camera view. In $K_{2D_{c}}[i]$, $c \in C$ denotes the set of all allocentric cameras, and $1\leq i \leq 21$ represents the 21 joints on the hand.

Given that not all positions are accurate, we employ RANSAC\cite{ransac} to filter out imprecise 2D positions. 
In every iteration of RANSAC, two 2D keypoint positions $K_{2D_{c_1}}[i]$ and $K_{2D_{c_2}}[i]$ are chosen from two randomly selected different camera views $c_1$ and $c_2$. Based on positions $K_{2D_{c_1}}[i]$ and $K_{2D_{c_2}}[i]$, we can calculate their corresponding 3D points $K_{3D}[i]_{<c_{1}, c_{2}>}$ in the world coordinate system via triangulation.
Subsequently, we project this 3D point onto camera planes and calculate the number of 2D keypoints within 30 pixels around the projected point. 
After all iterations, the 3D point $K_{3D}[i]_{<c_{1}, c_{2}>^*}$ with the highest number of surrounding 2D keypoints is selected as the 3D keypoint $K_{3D}[i]$. 
This process is defined as

\vspace{-0.5cm}
\begin{align}
    \label{eq:ransac}
    K_{3D}[i] = &\underset{K_{3D}[i]_{<c_{1}, c_{2}>}}{\arg \max } \sum_{c=1}^{12} \\
    \nonumber
    &around_{2D}(proj_{c}(K_{3D}[i]_{<c_{1}, c_{2}>}), K_{2D_{c}}[i], 30),
\end{align}
where $proj_{c}(\cdot)$ project $K_{3D}[i]$ onto camera $c$, and $around_{2D}(\cdot)$ calculates the distance between two points, outputting 1 if the distance is less than 30 pixels and 0 otherwise.
In the selected iteration, the 2D keypoints $K_{2D_{c}}[i]$ that are more than 30 pixels away from the projected point will be deemed invalid and will be excluded from the subsequent optimization stage as $valid_{c}[i] = around_{2D}(proj_{c}(K_{3D}[i]), K_{2D_{c}}[i], 30)$. 

\subsection{Details on Hand Pose Optimization}

We adopt MANO\cite{mano} to formulate a 3D hand mesh as $\Theta_{h} = \{\theta, \beta, t\}$, where $\theta \in \mathbb{R}^{48}$, $\beta \in \mathbb{R}^{10}$, and $ t \in \mathbb{R}^{3}$ represent hand pose, hand shape, and wrist position, respectively.
For each participant, the shape parameters $\beta$ are precomputed based on specially collected data with only two hands and remain fixed in the subsequent hand pose optimization process.
The MANO model maps $\Theta_{h}$ to a 3D hand mesh $\{J, V \} = MANO(\Theta_{h})$, where $J \in \mathbb{R}^{778 \times 3}$ and $J \in \mathbb{R}^{21 \times 3}$ represent vertices and joints on hand, respectively. 
we first fit a MANO model by minimizing the following loss function:

\vspace{-0.5cm}
\begin{align}
    \label{eq:hand optimization loss function stage 1}
    \hat{\Theta}_{h} = &\underset{\Theta_{h}}{\arg \min } \left(\lambda_{2D} \mathcal{L}_{2D} + \lambda_{3D} \mathcal{L}_{3D} + \lambda_{angle} \mathcal{L}_{angle} +  \right.  \\
                       \nonumber
                       & \left. \lambda_{tc} \mathcal{L}_{tc}\right),
\end{align}

where $\mathcal{L}_{2D}$ and $\mathcal{L}_{3D}$ encourages the MANO hand joints to align with the 2D and 3D keypoints, $\mathcal{L}_{angle}$ ensures a natural hand pose, and $\mathcal{L}_{tc}$ promotes temporal smoothness.
Utilizing object pose, we then refine hand pose by minimizing the following loss function:

\vspace{-0.5cm}
\begin{align}
    \label{eq:hand optimization loss function stage 2}
    \hat{\Theta}_{h} = &\underset{\Theta_{h}}{\arg \min } \left(\lambda_{2D} \mathcal{L}_{2D} + \lambda_{3D} \mathcal{L}_{3D} + \lambda_{angle} \mathcal{L}_{angle} +  \right.  \\[-0.5cm]
                       \nonumber
                       & \left. \lambda_{tc} \mathcal{L}_{tc} + \lambda_{p} \mathcal{L}_{p} + \lambda_{a} \mathcal{L}_{a}\right),
\end{align}

where $\mathcal{L}_{p}$ prevents hand-object interpenetration, and $\mathcal{L}_{a}$ encourages hand-object contact.

\textbf{2D joint loss $\mathcal{L}_{2D}$.}
The 2D joint loss term is defined as

\vspace{-0.5cm}
\begin{align}
    \label{eq:2d joint loss}
\mathcal{L}_{2D} = \sum_{c=1}^{12}\sum_{i=1}^{21}valid_{c}[i]\left \| proj_{c}(J[i]) - K_{2D_{c}}[i] \right \|^{2},
\end{align}

where $J[i]$ denotes the $i^{th}$ 3D hand joint position, the $proj_{c}(\cdot)$ operator projects it onto camera $c$, $K_{2D_{c}}[i]$ is the $i^{th}$ 2D keypoint position of hand in the camera view $c$, and $valid_{c}[i]$ which is determined in RANSAC indicates whether $K_{2D_{c}}[i]$ is a valid value.

\textbf{3D joint loss $\mathcal{L}_{3D}$.}
The 3D joint loss term is defined as
\begin{align}
    \label{eq:3d joint loss}
\mathcal{L}_{3D} = \sum_{i=1}^{21}\left \| J[i] - K_{3D}[i] \right \|^{2} 
\end{align}
 where $J_{i}$ denotes the $i^{th}$ 3D hand joint position and $K_{3D}[i]$ is the $i^{th}$ 3D keypoint position fused by 2D keypoint positions from 12 allocentric views in RANSAC. 2D joint loss $\mathcal{L}_{2D}$ and 3D joint loss $\mathcal{L}_{3D}$ provide the most direct supervision for hand pose, aligning the MANO hand with the positions of keypoints.

\textbf{Angle constraint loss $\mathcal{L}_{angle}$.}
The angle constraint loss term imposes restrictions on the permissible angles for the rotation of 15 joints, thus preventing undue distortion of the fingers and ensuring a natural hand pose.
In the MANO model, the hand pose parameter $\theta \in \mathbb{R}^{48}$, which can be conceptualized as $\theta \in \mathbb{R}^{16 \times 3}$, signifies 16 axis-angle representations. Among these, 1 axis-angle corresponds to the global rotation of the hand, while the remaining 15 axis-angle represent rotations of 15 joints on the hand. 
The angle constraint loss term is defined following \cite{zhou2016modelbased} as

\vspace{-0.5cm}
\begin{align}
    \label{eq:angle constraint loss}
\mathcal{L}_{angle} = \sum_{i=1}^{45} \max \left(\underline{\theta_{i}}-\theta[i], 0\right)+\max \left(\theta[i]-\overline{\theta_{i}}, 0\right),
\end{align}

where $\underline{\theta_{i}}$ and $\overline{\theta_{i}}$ denote the upper and lower bounds, respectively, for the $i^{th}$ joint angle parameter $\theta[i]$.

\textbf{Temporal consistency loss $\mathcal{L}_{tc}$.}
Due to noise in the data and the randomness in the output of the hand pose estimation model for each frame, the hand pose in the video may exhibit a noticeable degree of jitter.
While other loss terms are applied to individual frames, this loss term considers adjacent frames, helping to alleviate the jitter in the hand pose.
We draw inspiration from \cite{zhou2016modelbased} and define temporal consistency loss term as

\vspace{-0.5cm}
\begin{align}
    \label{eq:temporal consistency loss}
\mathcal{L}_{t c}=\sum_{i \in \mathcal{I}}\left(\left\|\Delta_{t}^{i}\right\|^{2}+\left\|\Delta_{\theta}^{i}-\Delta_{\theta}^{i-1}\right\|^{2}\right),
\end{align}

where $\Delta_{t}^{i} = t^{i} - t^{i-1}$ and $\Delta_{\theta}^{i} = \theta^{i} - \theta^{i-1}$. $\mathcal{I}$ represents the index number within the entire sequence, excluding the initial frame.

\textbf{Attraction loss $\mathcal{L}_{a}$.}
During the optimization process, there might be insufficient contact between the hand and the object.
The attraction loss term encourages the hands near the object to make sufficient contact with it and is defined as

\vspace{-0.5cm}
\begin{align}
    \label{eq:attraction loss}
\mathcal{L}_{a} = \sum_{i=1}^{778} 
around_{3D}(\mathrm{V}_{\mathrm{h}}[i], \mathrm{V}_{\mathrm{o}}\left[i^{*}\right], 0.01) \left\|\mathrm{V}_{\mathrm{h}}[i]-\mathrm{V}_{\mathrm{o}}\left[i^{*}\right]\right\|^{2},
\end{align}

where $\mathrm{V}_{\mathrm{h}}[i]$ is the $i^{th}$ vertex on hand mesh, $\mathrm{V}_{\mathrm{o}}\left[i^{*}\right]$ is the vertex on the object closest to $\mathrm{V}_{\mathrm{h}}[i]$, and $around_{3D}(\cdot)$ calculates the distance between two points, outputting 1 if the distance is less than 0.01 meter and 0 otherwise.
This loss term is computed twice: once between the right hand and the tool, and once between the left hand and the object.

\textbf{Penetration loss $\mathcal{L}_{p}$.}
During the optimization process, there is a possibility of interpenetration between the hand and the object. This is evidently unrealistic in real-world scenarios. Therefore, a loss term is introduced to mitigate such interpenetration.
Similar to \cite{hampali2020honnotate}, we define penetration loss term as

\vspace{-0.5cm}
\begin{align}
    \label{eq:penetration loss}
\mathcal{L}_{p} = \sum_{i=1}^{778} \max \left(-\mathbf{n}_{\mathrm{o}}\left(\mathrm{V}_{\mathrm{o}}\left[i^{*}\right]\right)^{T}\left(\mathrm{V}_{\mathrm{h}}[i]-\mathrm{V}_{\mathrm{o}}\left[i^{*}\right]\right), 0\right),
\end{align}

where the $\mathbf{n}_{\mathrm{o}}(\cdot)$ operator computes the normal for a vertex, and $\mathrm{V}_{\mathrm{h}}[i]$ represent the $i^{th}$ vertex on the hand mesh, and $\mathrm{V}_{\mathrm{o}}\left[i^{*}\right]$ denotes the vertex on the object closest to $\mathrm{V}_{\mathrm{h}}[i]$.
This loss term is computed twice: once between the right hand and the tool, and once between the left hand and the object.  

\section{Detailed Statistics on \dataset}

\begin{figure}[h!]
    \centering
    \includegraphics[width=1.0\linewidth]{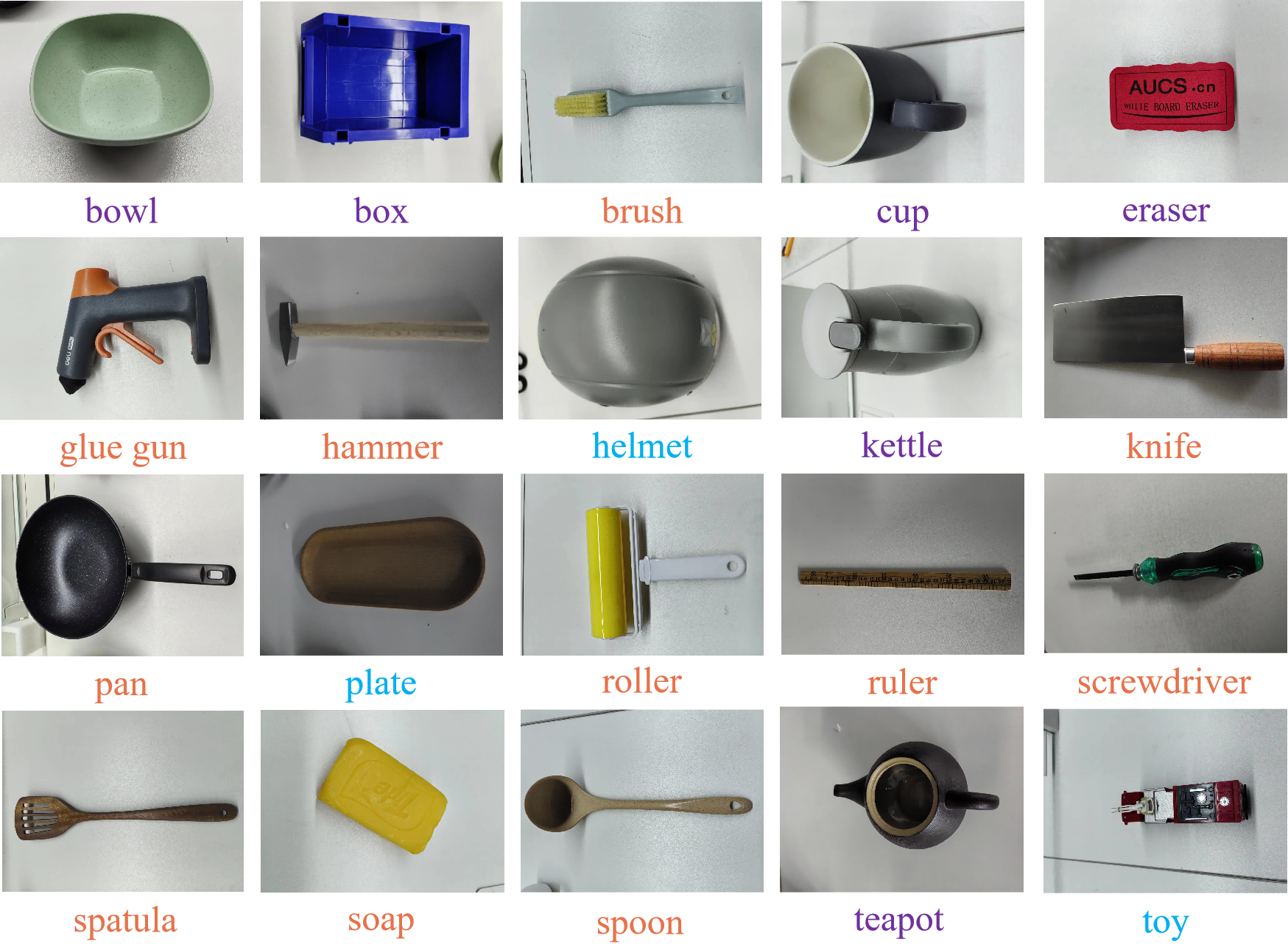}
    \vspace{-0.3cm}
    \caption{Visualization of 20 object categories in \dataset, including 17 tool categories (shown in purple and brown) and 9 target object categories (shown in purple and blue).}
    \vspace{-0.3cm}
    \label{fig:object_categories}
\end{figure}

\begin{figure}[h!]
    \centering
    \includegraphics[width=0.9\linewidth]{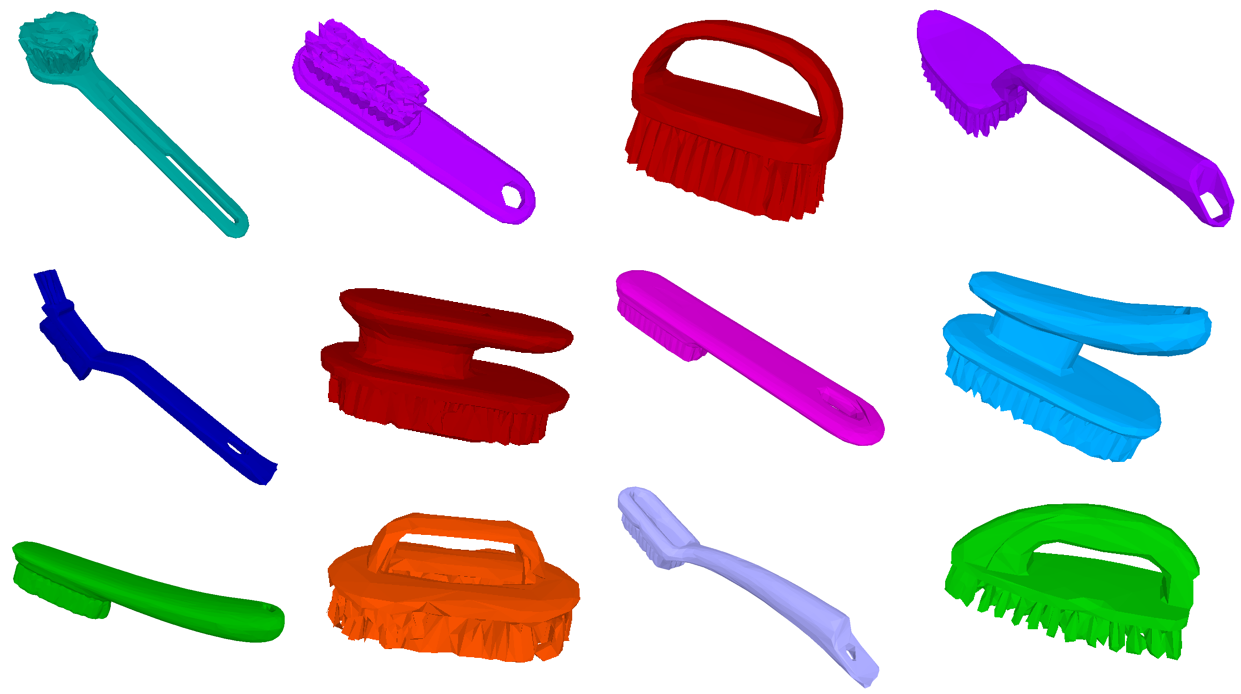}
    \vspace{-0.3cm}
    \caption{Visualization of 12 brushes in \dataset.}
    \vspace{-0.5cm}
    \label{fig:object_geometries}
\end{figure}

\textbf{Object diversities.} Figure \ref{fig:object_categories} shows the 20 object categories in our dataset. The categories are chosen from everyday hand-object interaction scenarios, and each object has a proper scale that can be manipulated stably by a single hand. Among these categories, 17 categories are utilized as tools during interaction, and 9 categories are treated as target objects. Figure \ref{fig:object_geometries} illustrates 12 object instances from the \textit{brush} category, indicating the diversity of object geometries.

\begin{figure}[h!]
    \centering
    \includegraphics[width=1.0\linewidth]{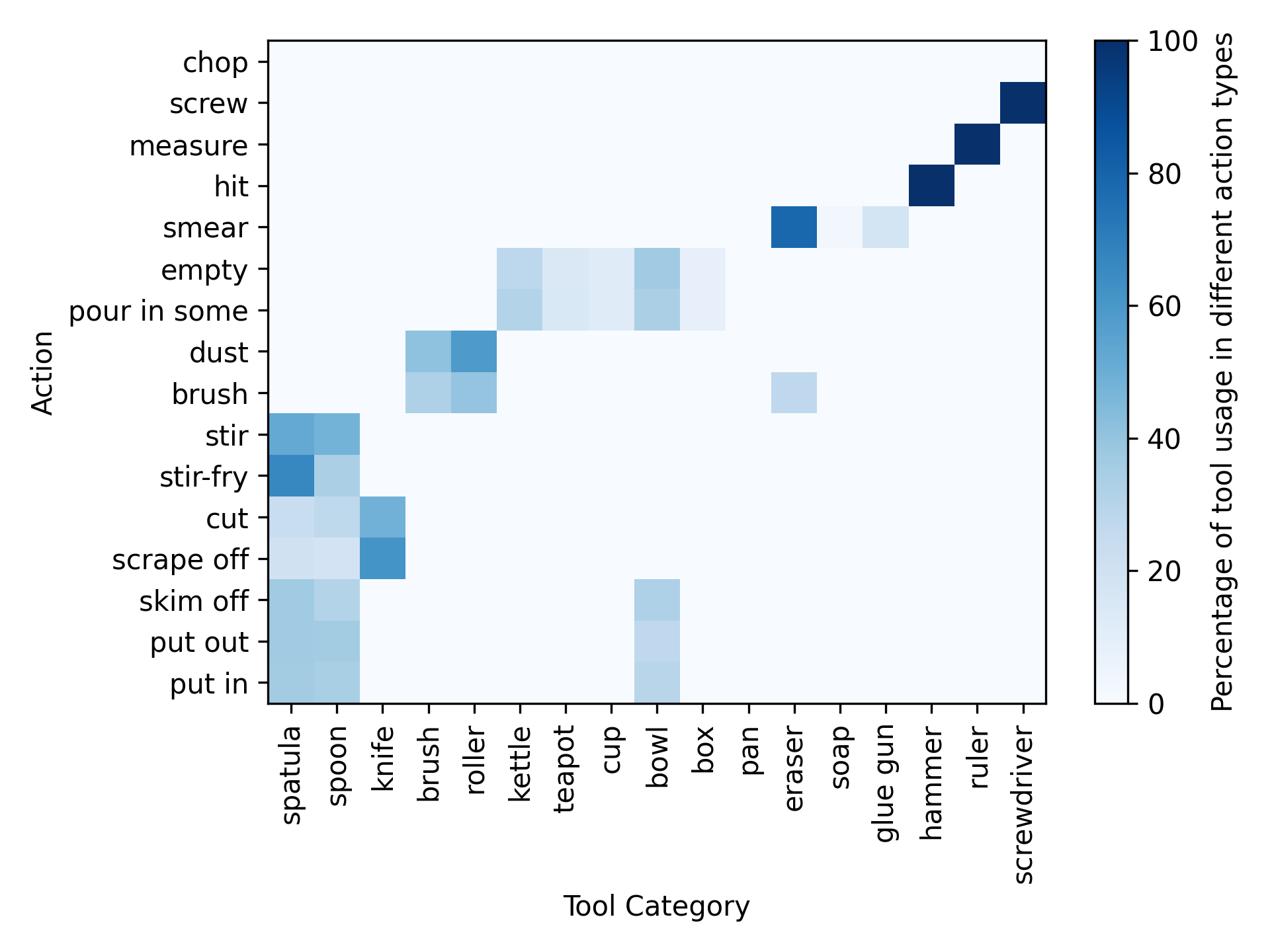}
    \vspace{-0.5cm}
    \caption{Percentage of tool usage for each action type.}
    \vspace{-0.5cm}
    \label{fig:tool_action_distribution}
\end{figure}

\begin{figure}[h!]
    \centering
    \includegraphics[width=1.0\linewidth]{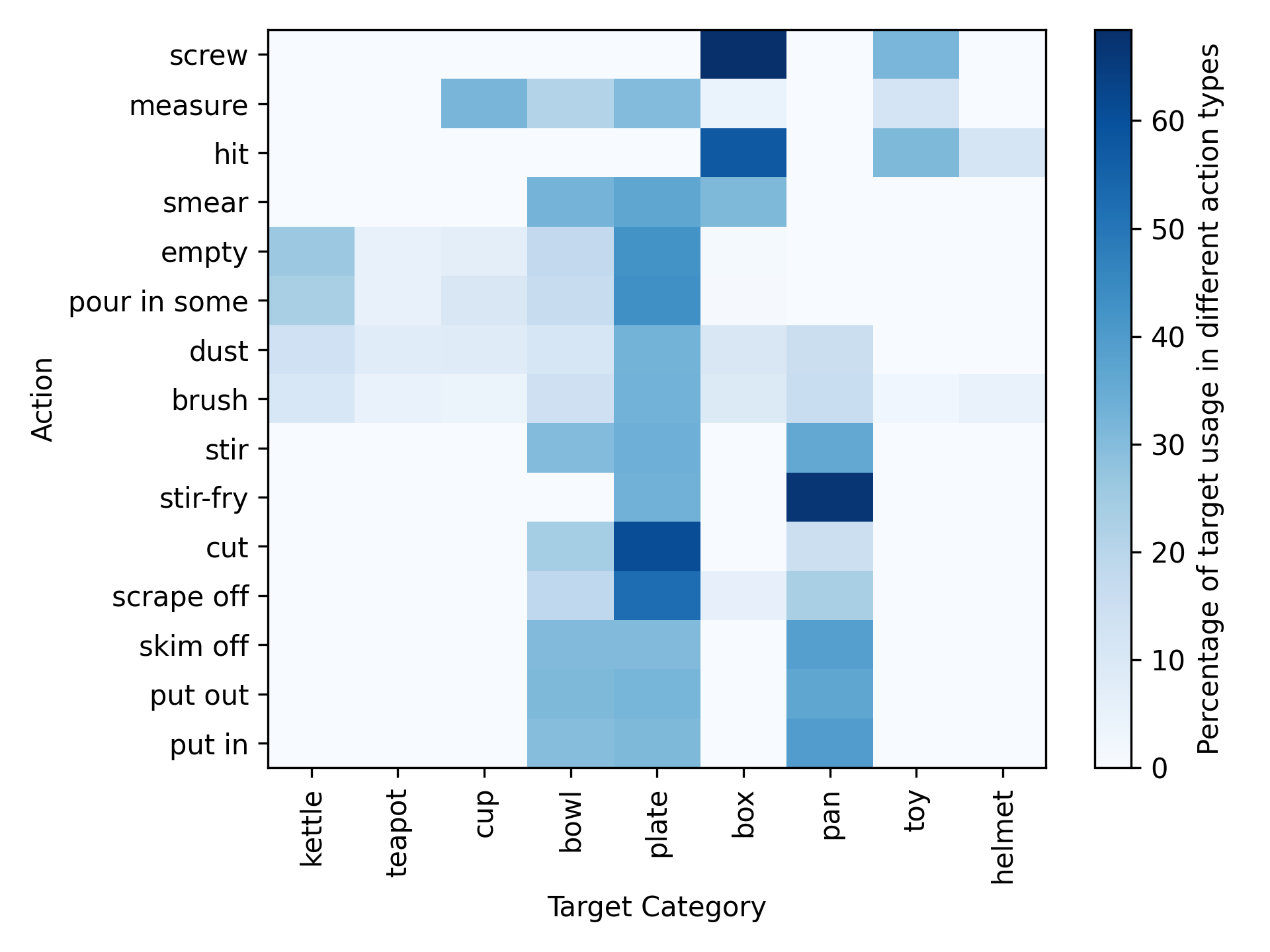}
    \vspace{-0.5cm}
    \caption{Percentage of target object usage for each action type.}
    \vspace{-0.3cm}
    \label{fig:target_action_distribution}
\end{figure}

\textbf{Interaction diversities.} As a knowledge base supporting generalizable studies on novel tool-action-object triplets, \dataset includes using different tools and target objects to perform the same action types. Figure \ref{fig:tool_action_distribution} and \ref{fig:target_action_distribution} show percentages of tool and target object usages in different action types, respectively. All 15 action types involve interaction demonstrations from various kinds of target object categories, while 12 out of them are performed by multiple tool categories.

\textbf{Motion speed.} Table \ref{tab:motion_speed} shows statistics on the speed of hand-object motions from different action types. $v_r$, $v_j$, $v$, and $\omega$ represent the velocity of the hand wrist, the average velocity of the MANO hand joints, the linear velocity of the object, and the angular velocity of the object, respectively. Compared to target objects, tools always dominate in interaction and have a significantly fast motion speed (16.6 cm/s and 71.3 $^\circ$/s on average), indicating the difficulties of forecasting and synthesizing their motions. Since all manipulation behaviors are performed by right-handed individuals, the right hand commonly controls the tool and thus consistently moves faster than the left hand among different action types.

\begin{table*}
    \centering
    \small
    \begin{tabular}{|c|cc|cc|cc|cc|}
        \hline
        \multirow{2}{*}{Action} & \multicolumn{2}{c|}{Right hand} & \multicolumn{2}{c|}{Left hand} & \multicolumn{2}{c|}{Tool} & \multicolumn{2}{c|}{Target object} \\
        \cline{2-9}
        & $v_r$(cm/s) & $v_j$(cm/s) & $v_r$(cm/s) & $v_j$(cm/s) & $v$(cm/s) & $\omega$($^\circ$/s) & $v$(cm/s) & $\omega$($^\circ$/s) \\
        \hline
        Put in & 17.2($\pm$11.5) & 20.0($\pm$13.2) & 9.4($\pm$10.5) & 10.4($\pm$11.8) & 18.9($\pm$20.9) & 88.1($\pm$116.3) & 3.7($\pm$6.6) & 14.1($\pm$35.3) \\
        \hline
        Put out & 17.9($\pm$11.6) & 20.5($\pm$13.0) & 10.5($\pm$10.5) & 11.7($\pm$11.7) & 18.7($\pm$20.9) & 90.8($\pm$129.1) & 4.4($\pm$7.5) & 18.2($\pm$67.0) \\
        \hline
        Skim off & 15.4($\pm$10.7) & 18.0($\pm$12.5) & 9.4($\pm$10.3) & 10.2($\pm$11.4) & 17.1($\pm$19.9) & 70.8($\pm$98.9) & 3.7($\pm$6.4) & 15.0($\pm$41.5) \\
        \hline
        Scrape off & 13.4($\pm$10.4) & 15.4($\pm$11.6) & 9.6($\pm$10.0) & 10.7($\pm$11.6) & 16.7($\pm$19.0) & 69.5($\pm$98.9) & 4.0($\pm$7.4) & 19.0($\pm$47.8) \\
        \hline
        Cut & 13.7($\pm$11.5) & 15.5($\pm$13.0) & 8.4($\pm$9.8) & 9.5($\pm$11.3) & 15.4($\pm$17.9) & 70.1($\pm$115.3) & 3.1($\pm$6.1) & 13.3($\pm$32.5) \\
        \hline
        Stir-fry & 17.0($\pm$12.2) & 19.3($\pm$13.3) & 10.7($\pm$9.7) & 12.1($\pm$11.7) & 21.4($\pm$20.5) & 77.0($\pm$88.2) & 8.2($\pm$15.5) & 23.1($\pm$63.7) \\
        \hline
        Stir & 16.8($\pm$12.6) & 18.9($\pm$13.7) & 9.0($\pm$10.0) & 10.3($\pm$11.2) & 20.1($\pm$20.3) & 74.0($\pm$114.6) & 4.6($\pm$7.4) & 15.6($\pm$29.5) \\
        \hline
        Brush & 15.3($\pm$12.2) & 17.7($\pm$14.5) & 11.0($\pm$10.4) & 12.2($\pm$11.9) & 19.3($\pm$22.3) & 71.5($\pm$167.6) & 8.1($\pm$11.4) & 33.9($\pm$55.6) \\
        \hline
        Dust & 13.9($\pm$9.5) & 17.2($\pm$11.4) & 10.8($\pm$10.4) & 12.4($\pm$12.1) & 19.6($\pm$21.9) & 88.5($\pm$121.1) & 7.1($\pm$10.6) & 31.6($\pm$49.7) \\
        \hline
        Pour in some & 15.0($\pm$13.0) & 16.3($\pm$14.2) & 9.0($\pm$10.8) & 10.0($\pm$12.7) & 8.8($\pm$13.2) & 39.4($\pm$77.9) & 2.8($\pm$5.2) & 11.1($\pm$27.4) \\
        \hline
        Empty & 14.9($\pm$13.6) & 16.1($\pm$14.7) & 8.5($\pm$10.6) & 9.4($\pm$12.5) & 10.2($\pm$14.2) & 45.5($\pm$83.49) & 2.9($\pm$5.4) & 12.3($\pm$40.0) \\
        \hline
        Smear & 12.4($\pm$12.5) & 15.8($\pm$15.2) & 9.9($\pm$11.4) & 11.4($\pm$12.3) & 16.5($\pm$19.5) & 60.1($\pm$161.6) & 7.4($\pm$10.8) & 32.7($\pm$59.5) \\
        \hline
        Hit & 14.1($\pm$11.1) & 17.4($\pm$13.1) & 9.3($\pm$9.6) & 9.9($\pm$9.9) & 18.7($\pm$23.9) & 67.6($\pm$114.1) & 5.7($\pm$10.1) & 21.9($\pm$40.0) \\
        \hline
        Measure & 15.3($\pm$11.3) & 16.0($\pm$12.4) & 15.9($\pm$13.8) & 17.8($\pm$16.7) & 12.0($\pm$14.8) & 56.9($\pm$102.0) & 2.0($\pm$5.6) & 10.6($\pm$24.2) \\
        \hline
        Screw & 12.5($\pm$11.7) & 15.7($\pm$13.0) & 10.8($\pm$12.2) & 11.7($\pm$13.0) & 11.6($\pm$18.5) & 182.5($\pm$278.0) & 4.8($\pm$8.7) & 20.7($\pm$37.1) \\
        \hline
        \textit{Overall} & 15.0($\pm$11.7) & 17.2($\pm$13.4) & 9.9($\pm$10.6) & 11.1($\pm$12.1) & 16.6($\pm$19.9) & 71.3($\pm$125.9) & 5.0($\pm$8.8) & 20.9($\pm$46.9) \\
        \hline
    \end{tabular}
    \vspace{-0.3cm}
    \caption{Average hand and object motion speed for each specific action type.}
    \vspace{-0.3cm}
    \label{tab:motion_speed}
\end{table*}

\begin{figure}[h!]
    \centering
    \includegraphics[width=1.0\linewidth]{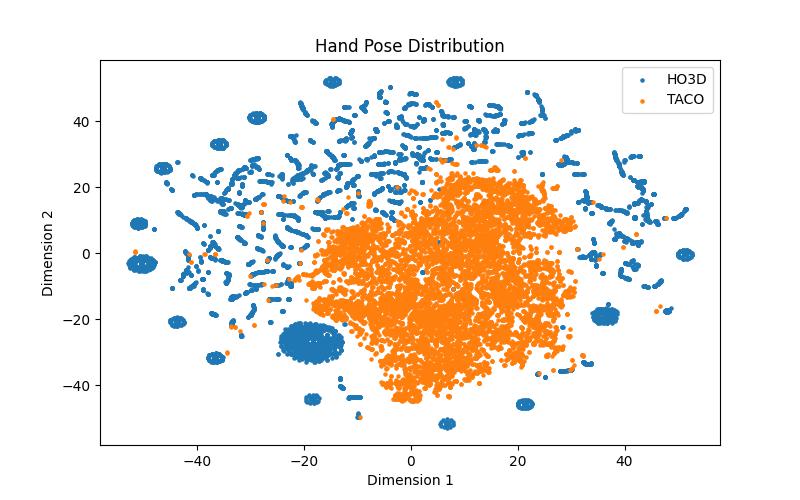}
    \vspace{-0.5cm}
    \caption{T-SNE visualization of hand poses from \dataset and HO3D.}
    \vspace{-0.5cm}
    \label{fig:hand_pose_TSNE}
\end{figure}

\textbf{Hand pose distribution.} Figure \ref{fig:hand_pose_TSNE} illustrates the T-SNE visualization of hand poses from \dataset and HO3D\cite{hampali2020honnotate}. The distribution of hand poses from \dataset mostly differs from that of HO3D due to the different human behaviors.

\section{Details on Marker Removal Evaluation}

\textbf{Data processing.} For each image from the raw videos and the marker-removed ones, we first render the spheres on the image to obtain their 2D mask and crop an image patch with the boundary the same as the mask. We then scale the image patch in equal proportions and place it at the center of a 512x512 image with black background color. Finally, a Gaussian kernel with $\sigma$=$1.0$ is utilized to augment the 512x512 image as the network input.

\textbf{Network training.} For an image input $I \in \mathbb{R}^{512\times 512\times 3}$, a U-Net\cite{U-Net} is used to estimate the heatmap $H\in \mathbb{R}^{512\times 512}$ that indicates the probability that each pixel belongs to the inpainted image regions. The loss function is the mean-square error comparing the estimated heatmaps against the ground truth ones. The network is trained by an Adam~\cite{kingma2014adam} optimizer with a learning rate of 5e-4.

\section{Details on Evaluation Metrics}

\textbf{Evaluating compositional action recognition.} Following existing action recognition work\cite{yang2023aim,radevski2021revisiting}, we use Top-1 Accuracy and Top-5 Accuracy to evaluate whether the ground truth action label appears in the top-1 or top-5 predictions with the highest probabilities presented by the method.

\textbf{Evaluating generalizable hand-object motion forecasting.} As a prediction task, motion forecasting approaches are assessed by measuring differences between predictions and ground truths. Following the line of human-object motion forecasting studies\cite{CAHMP,adeli2021tripod,HO-GCN,InterDiff}, we represent a hand as a 3D skeleton $J \in \mathbb{R}^{21\times3}$ with 21 joints and use Mean Per Joint Position Error $J_e=\frac{1}{21M} \sum_{k=1}^{M} \sum_{i=1}^{21} \Vert \hat{J}_{k,i} - \bar{J}_{k,i} \Vert_2$ to measure hand predictions, where M is the number of predicted frames, $\hat{J}$ is hand skeleton predictions, and $\bar{J}$ is ground truth values. Since objects are rigid bodies, the translation error $T_e$ and rotation error $R_e$ are defined as:

\vspace{-0.5cm}
\begin{equation}
\begin{split}
& T_e = \frac{1}{M} \sum_{k=1}^{M} \Vert \hat{t}_k - \bar{t}_k \Vert_2, \\
& R_e = \frac{1}{M} \sum_{k=1}^{M} \arccos{\frac{\mathbf{Tr}(\hat{R}_k^T \bar{R}_k)-1}{2}},
\end{split}
\end{equation}

where $\hat{t} \in \mathbb{R}^3$ and $\hat{R} \in \mathbb{R}^{3\times3}$ are predicted object translation vectors and rotation matrices, $\bar{t}$ and $\bar{R}$ are ground-truth ones, and $\mathbf{Tr}$ denotes the trace of a matrix.

\begin{figure}[h!]
    \centering
    \includegraphics[width=1.0\linewidth]{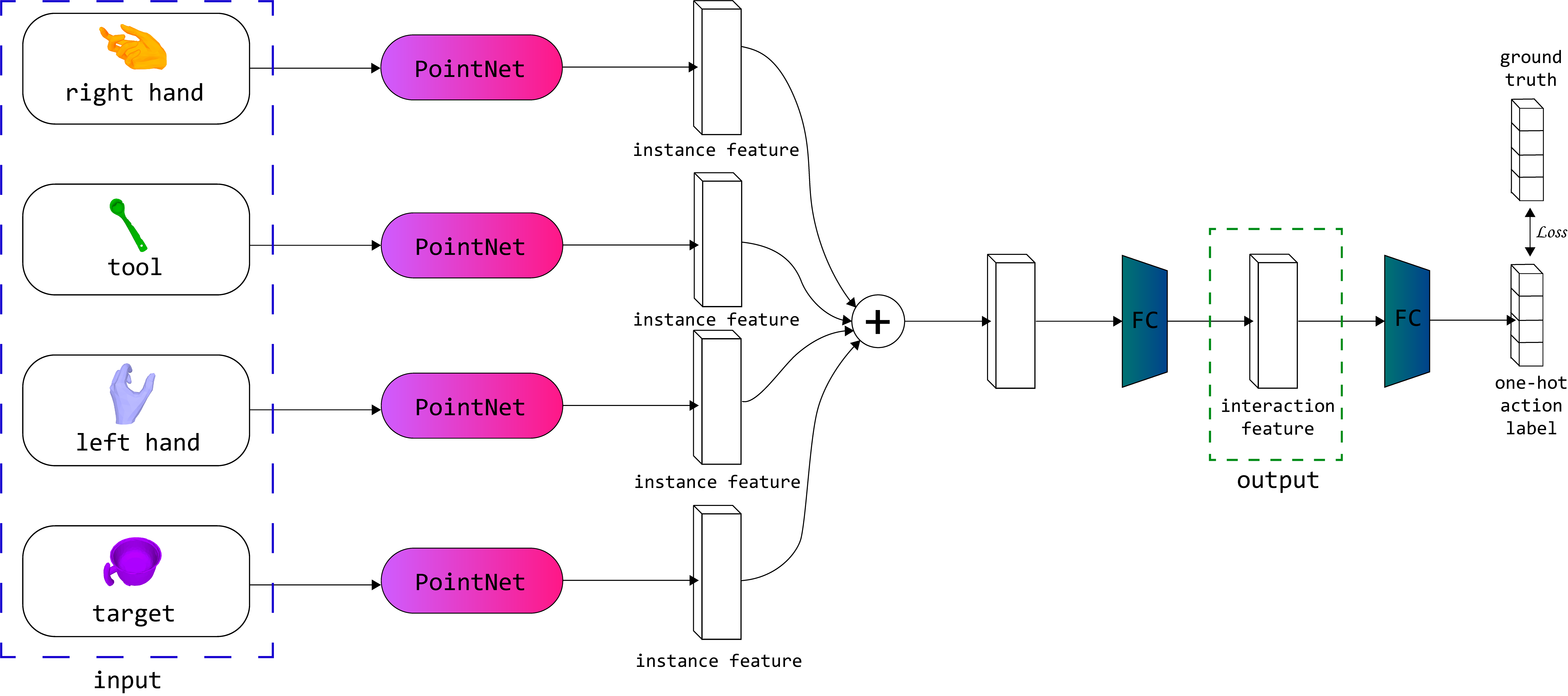}
    \vspace{-0.5cm}
    \caption{Network structure of our interaction feature extractor. Given vertices of hand-object meshes, the network first utilizes PointNet\cite{qi2017pointnet} to encode vertices of each mesh to a 128-dimensional feature, respectively, and then adds the four features together and acquires the interaction feature via a fully connected layer.}
    \vspace{-0.3cm}
    \label{fig:fid}
\end{figure}

\textbf{Evaluating cooperative grasp synthesis.} As a generative task, the benchmark should examine the physical plausibility and reality of synthesized hand meshes. For assessing physical plausibility, the contact ratio (\textit{Con. R}) indicates the proportion of results that are in contact with the tool, while the interpenetration volume (\textit{Pen. V}) denotes the average volume that is occupied by both the generated hand and the tool and is computed by voxelizing hand-object meshes to 1mm cubes and counting the intersecting ones. The collision ratio (\textit{Col. R}) examines conflicts between generated hands and the environment, computing the probability of results penetrating the target object and the left hand. To evaluate whether results are realistic, we first present an interaction feature extractor (Figure \ref{fig:fid}) that encodes hand-object vertices to a 64-dimensional feature $f$ and obtain ground truth feature distribution $\bar{D}=\{\bar{f}_i\}$ by applying it to real interaction snapshots. We then replace the vertices of the right-hand mesh with those from synthesized ones and obtain another feature distribution $\hat{D}=\{\hat{f}_i\}$. Finally, the Fr{\'e}chet Inception Distance (FID) score (\textit{FID}) is computed on $\hat{D}$ and $\bar{D}$ measuring the dissimilarity between them. The interaction feature extractor is supervised-trained by decoding the one-hot action label from $f$ via a fully connected layer.

\section{Baseline Designs for Interactive Grasp Synthesis}

\begin{figure}[h!]
    \centering
    \includegraphics[width=0.98\linewidth]{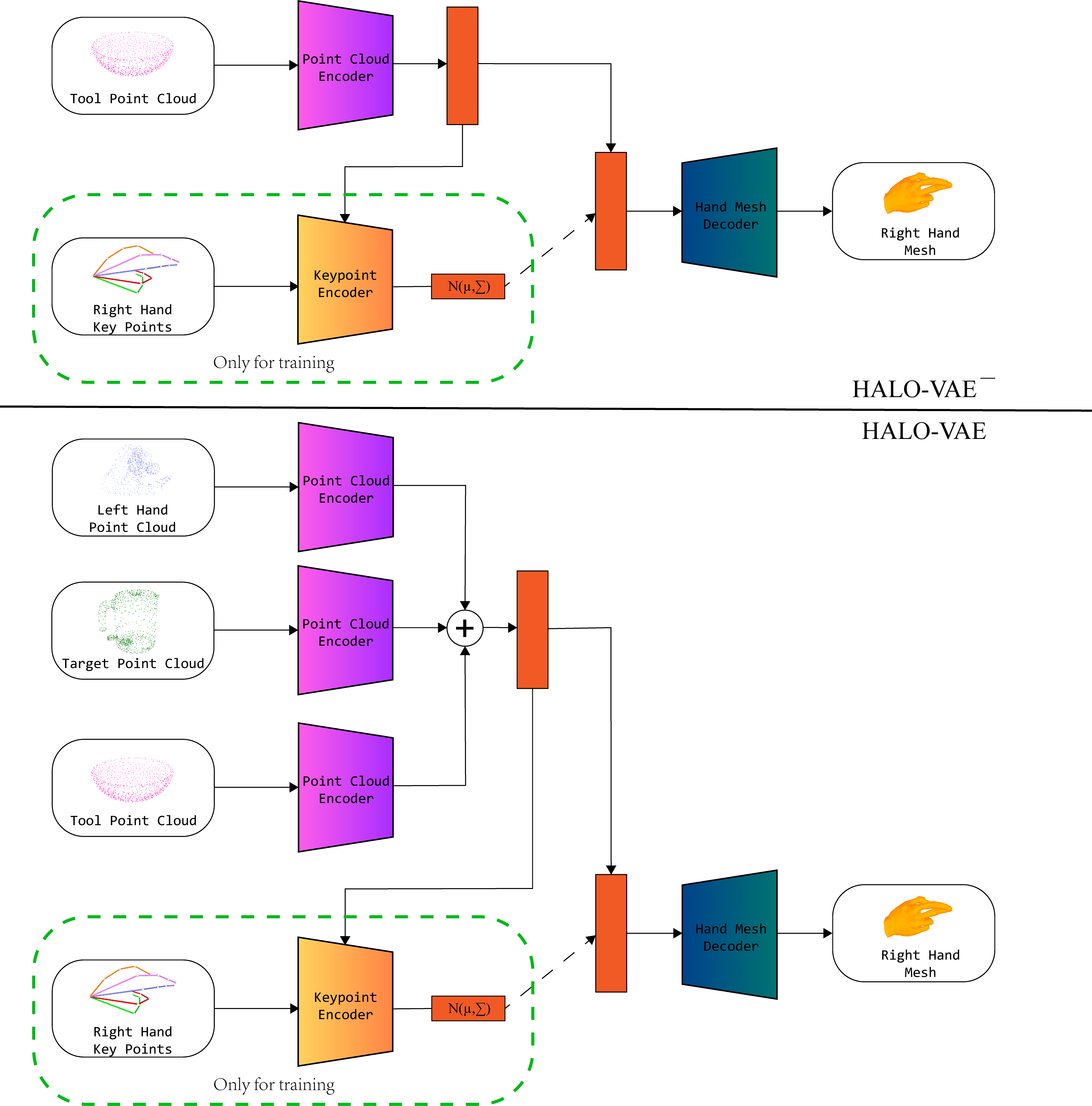}
    \vspace{-0.2cm}
    \caption{Comparison of HALO-VAE$^-$ and our modified HALO-VAE\cite{HALO}.}
    \vspace{-0.6cm}
    \label{fig:halo_vae_structure}
\end{figure}

We modify baseline approaches\cite{ContactGen,HALO} to integrate the interaction environment (the left hand and the target object) into the network structure. We directly regard the interaction environment as additional conditions for CVAE, and apply existing point cloud encoders to transfer its point clouds to feature vectors. Figure \ref{fig:halo_vae_structure} compares our modified HALO-VAE\cite{HALO} structure with HALO-VAE$^-$.

\section{Qualitative Results on Hand-object Motion Forecasting}

\begin{figure}[h!]
    \centering
    \includegraphics[width=1.0\linewidth]{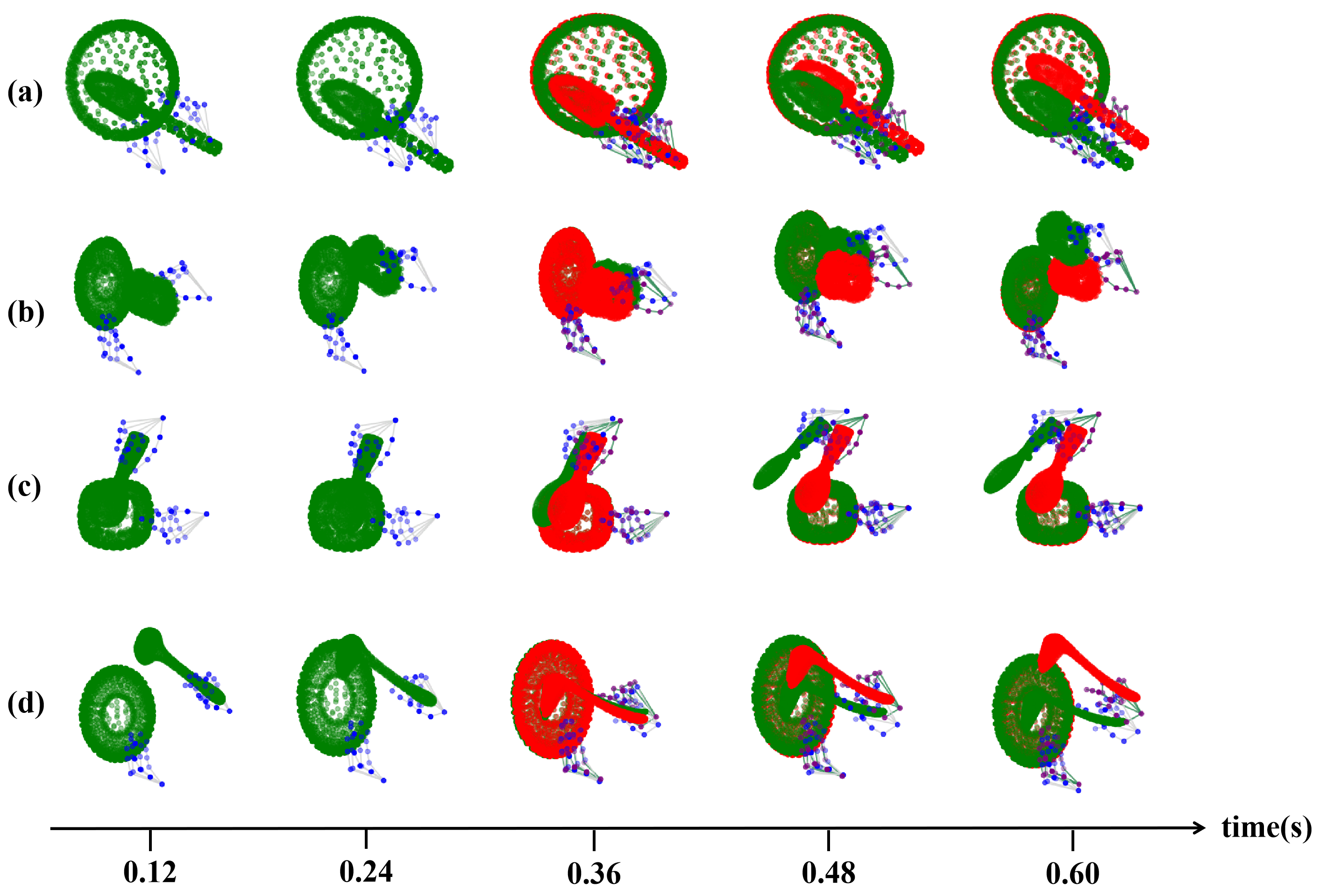}
    \vspace{-0.3cm}
    \caption{Qualitative results on hand-object motions predicted by CAHMP\cite{CAHMP}. The green and blue points denote the ground truth, while the red and purple ones indicate the predicted motions. We show five frames at 0.12, 0.24, 0.36, 0.48, and 0.60s.}
    \vspace{-0.5cm}
    \label{fig:vis_motion_forecasting}
\end{figure}

Figure \ref{fig:vis_motion_forecasting} shows the qualitative results of CAHMP\cite{CAHMP}. Although CAHMP achieves the best performance among the four baseline methods, it commonly fails to forecast fast movements (Figure \ref{fig:vis_motion_forecasting} (a),(b)) from the right hand and the tool, and encounters difficulty understanding human interaction intentions (Figure \ref{fig:vis_motion_forecasting} (c),(d)). Please see our supplementary video for more visualizations.

\section{\dataset Visualization}

\begin{figure}[H]
    \centering
    \includegraphics[width=1.0\linewidth]{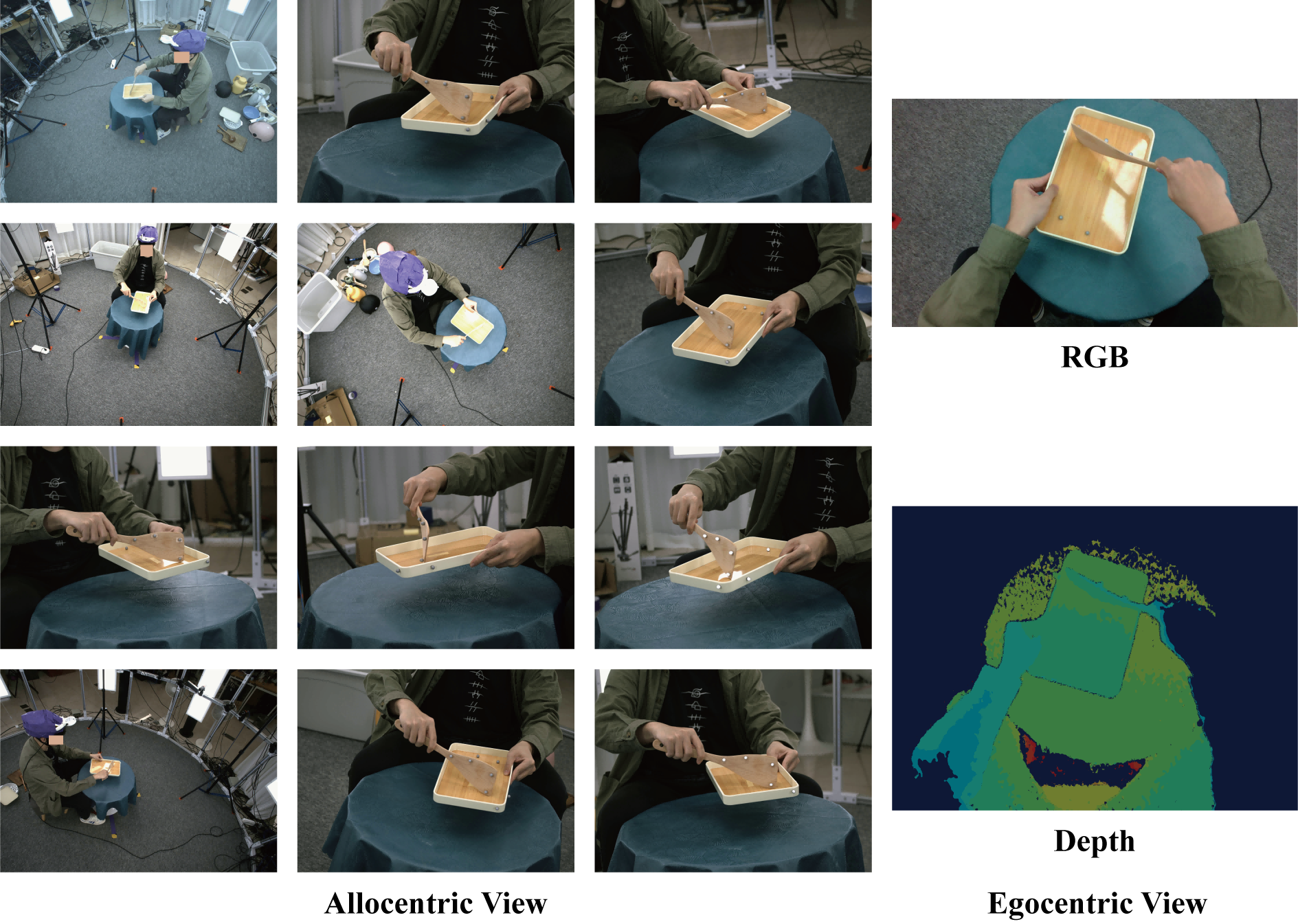}
    \vspace{-0.3cm}
    \caption{Visualization of allocentric and egocentric camera views. Our system involves 12 allocentric RGB cameras and one egocentric RGBD sensor.}
    \vspace{-0.5cm}
    \label{fig:camera_visualization}
\end{figure}

\begin{figure*}[h!]
    \centering
    \includegraphics[width=1.0\linewidth]{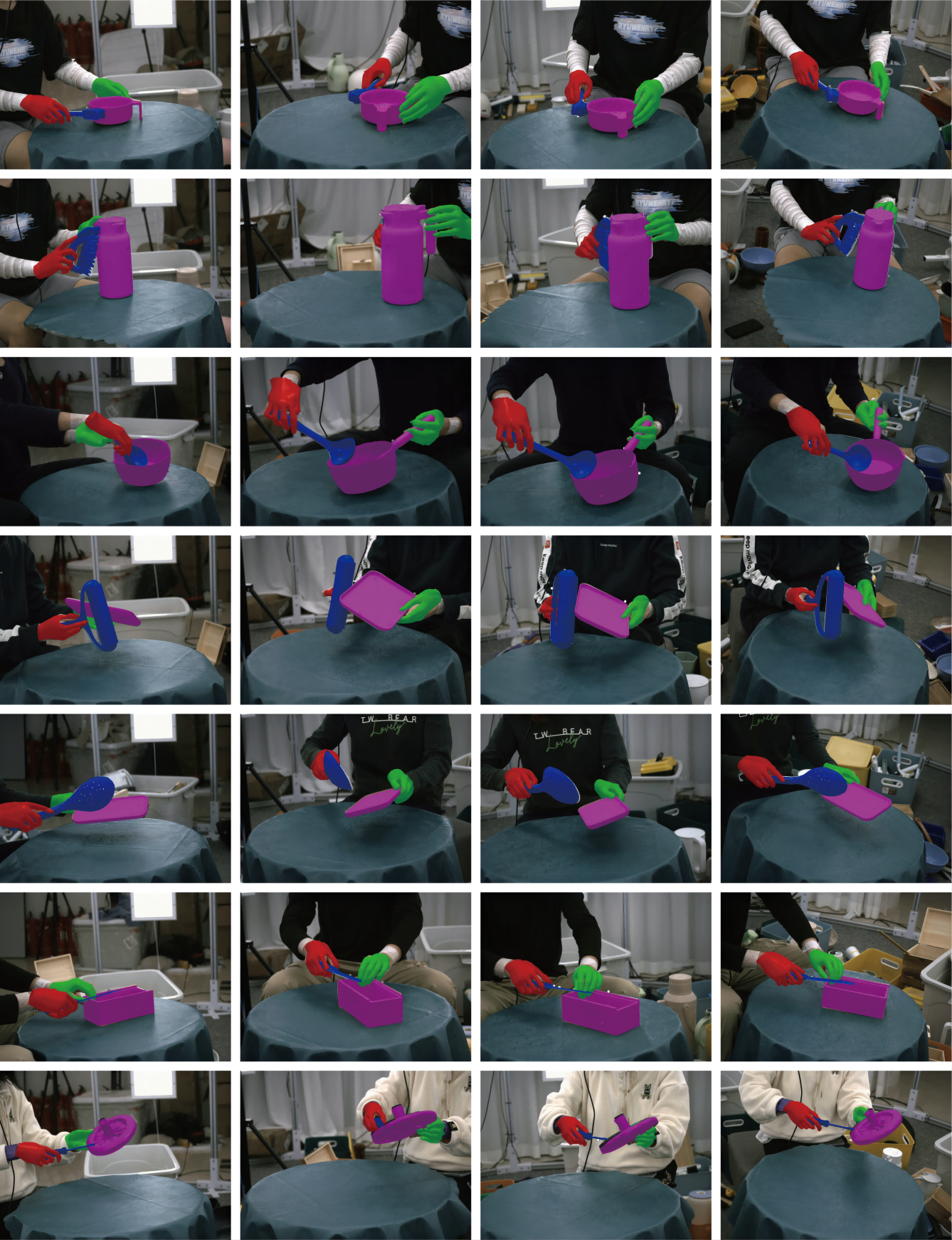}
    \vspace{-0.3cm}
    \caption{Visualization of hand-object meshes. We overlay the original color frames with rendered hand-object meshes.}
    \vspace{-0.5cm}
    \label{fig:HO_mesh_visualization}
\end{figure*}

\begin{figure*}[h!]
    \centering
    \includegraphics[width=1.0\linewidth]{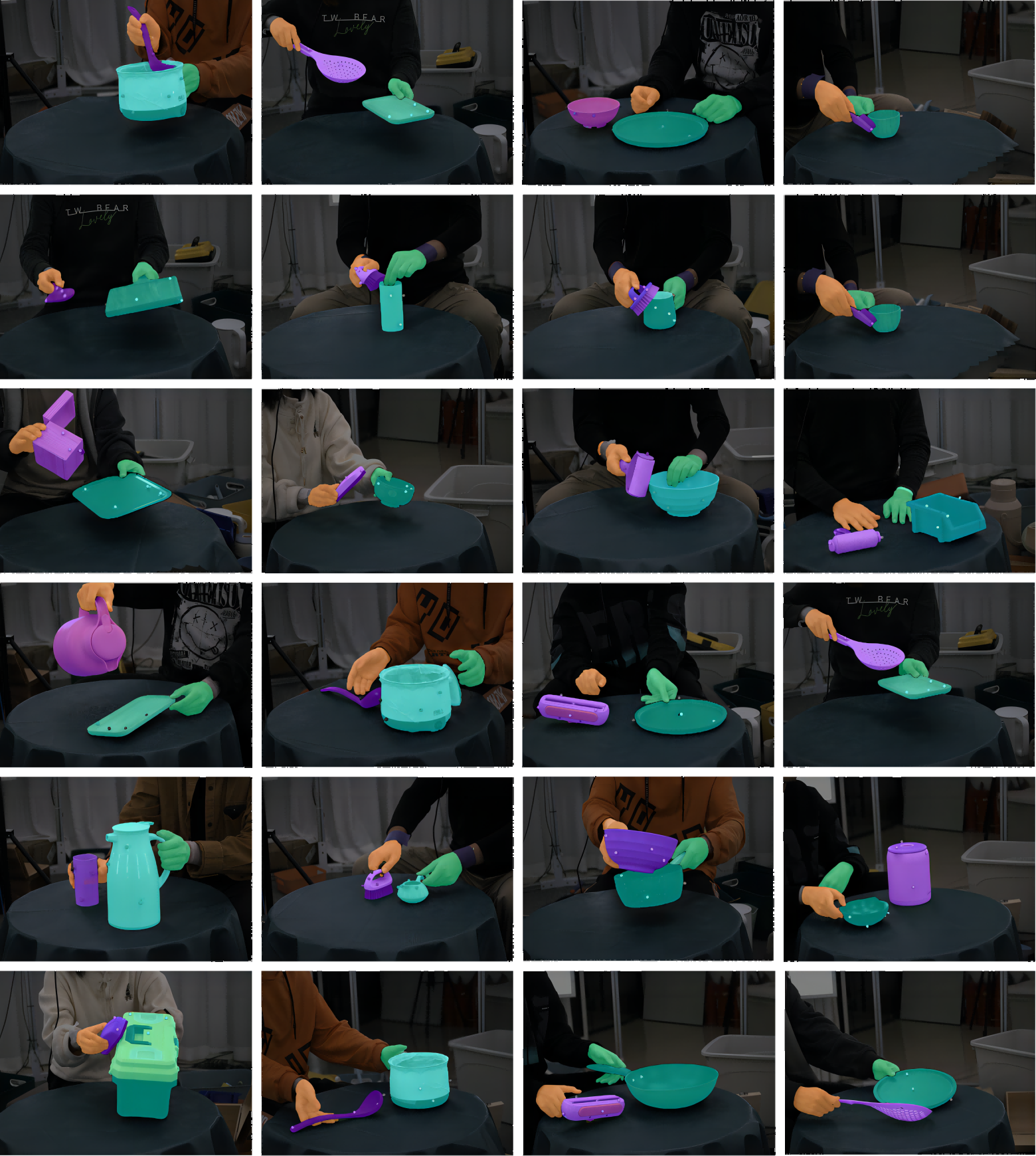}
    \vspace{-0.3cm}
    \caption{Visualization of automatic 2D hand-object segmentation.}
    \vspace{-0.5cm}
    \label{fig:2Dmask_visualization}
\end{figure*}

\begin{figure*}[h!]
    \centering
    \includegraphics[width=1.0\linewidth]{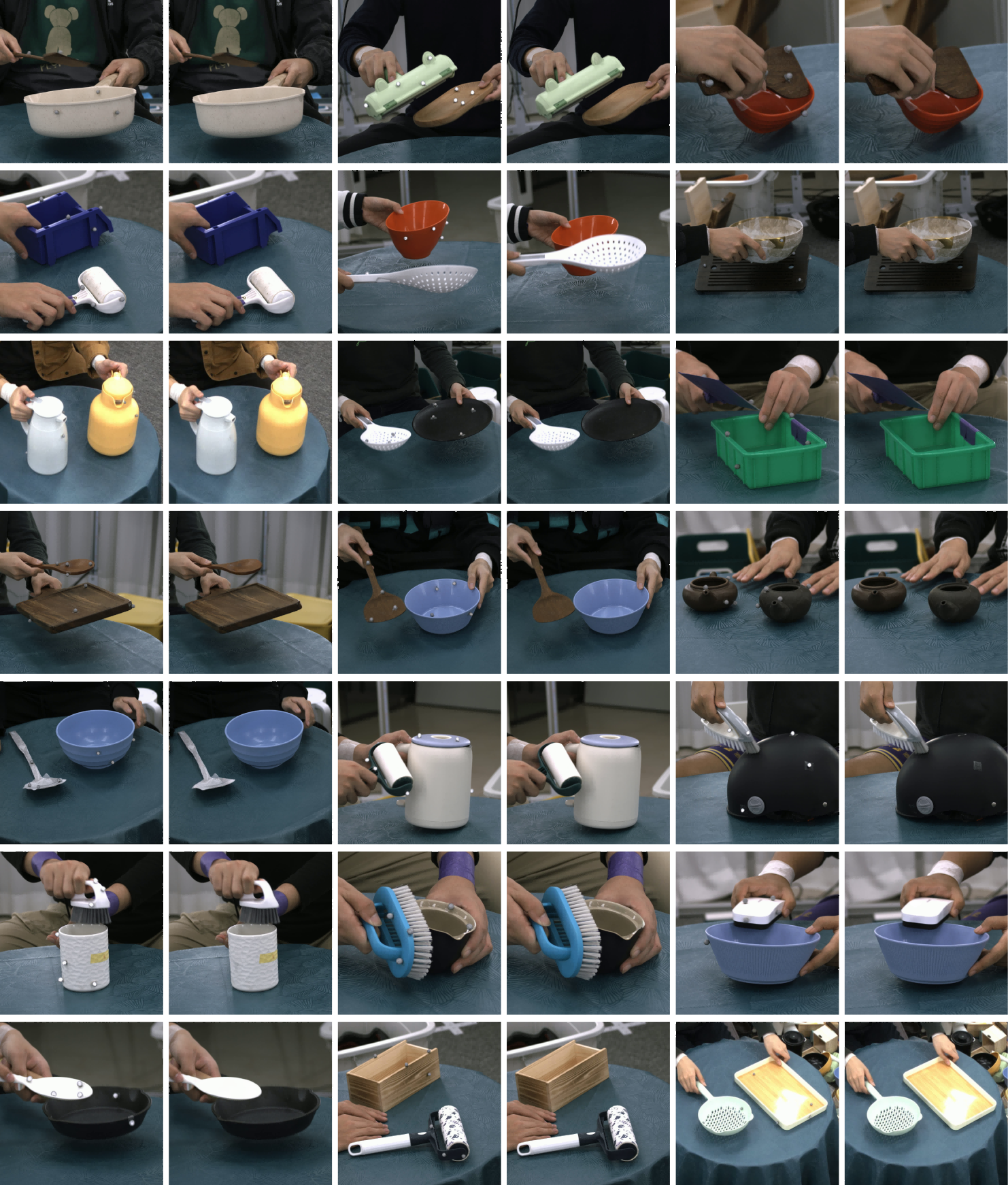}
    \vspace{-0.3cm}
    \caption{Visualization of original and marker-removed image patches.}
    \vspace{-0.5cm}
    \label{fig:marker_removal_visualization}
\end{figure*}

Figure \ref{fig:camera_visualization} shows our 12 RGB frames from all third-person views and the RGB and depth images from our egocentric camera. Figures \ref{fig:HO_mesh_visualization}, \ref{fig:2Dmask_visualization}, and \ref{fig:marker_removal_visualization} exhibit some examples of our hand-object meshes, hand-object segmentation, and marker-removed image patches, respectively. Please see our supplementary video for more data visualizations.

\end{document}